\documentclass[12pt]{article}
\usepackage[margin=1in]{geometry}
\usepackage[T1]{fontenc}
\usepackage[utf8]{inputenc}
\usepackage{lmodern}
\usepackage{microtype}
\usepackage{amsmath,amssymb,amsthm,mathtools,bm}
\usepackage{booktabs}
\usepackage{graphicx}
\usepackage{adjustbox}
\usepackage{placeins}
\usepackage{comment}
\graphicspath{{integral-curve-20261012/figures/}{figures/}}
\usepackage{longtable}
\usepackage{enumitem}
\usepackage{algorithm}
\usepackage{algpseudocode}
\usepackage[round,authoryear]{natbib}
\usepackage{xcolor}
\usepackage{setspace}
\usepackage[hidelinks]{hyperref}
\numberwithin{equation}{section}
\newtheorem{theorem}{Theorem}[section]

\newtheorem{corollary}[theorem]{Corollary}

\newtheorem{assumption}[theorem]{Assumption}
\theoremstyle{definition}
\newtheorem{remark}[theorem]{Remark}

\newcommand{\M}{\mathcal M}
\newcommand{\R}{\mathbb R}
\newcommand{\E}{\mathbb E}
\newcommand{\PP}{\mathbb P}

\newcommand{\cP}{\mathcal P}
\newcommand{\Exp}{\operatorname{Exp}}
\newcommand{\Log}{\operatorname{Log}}

\newcommand{\Cov}{\operatorname{Cov}}
\newcommand{\tr}{\operatorname{tr}}

\newcommand{\dd}{\,\mathrm d}

\newcommand{\eps}{\varepsilon}
\newcommand{\wh}{\widehat}

\newcommand{\norm}[1]{\left\lVert #1\right\rVert}
\newcommand{\abs}[1]{\left\lvert #1\right\rvert}

\title{Sampling Design and Confidence Tubes for Integral Curves on Riemannian Manifolds}
\author{Xiaotian Chang\thanks{Division of Mathematical Sciences, Nanyang Technological University}\hspace{0.4em},
Qirui Hu \thanks{\hangindent=1.8em\hangafter=1 School of Statistics and Data Science, Shanghai University of Finance and Economics.\\
Department of Mathematics, Ruhr University Bochum}\thanks{Corresponding author: huqirui@mail.shufe.edu.cn}\hspace{0.6em}and
Cyrus Mostajeran\footnotemark[1]\hspace{0.4em}\textsuperscript{,}\hspace{0.1em}\hspace{0.4em}}
\date{}

\begin{document}
\maketitle

\begin{abstract}
We estimate the integral curve of an unknown vector field from noisy tangent vector measurements on a known compact Riemannian manifold. We consider two sampling settings: independent observations at random locations that have already been collected, and a fixed response budget that allows subsequent locations to be chosen using pilot information. Under suitable regularity assumptions, the passive kernel estimator has curve error of order $n^{-2/(d+3)}$ in probability, with sample size $n$ and manifold dimension $d$. Its functional central limit theorem has a limit characterized by a linear stochastic differential equation. For active sampling, an independent pilot directs one new batch of measurements to neighborhoods of the target curve. We further prove that this concentration yields expected maximum curve error of order $N^{-1/2}$, including the pilot in the total response budget $N$. Both methods provide simultaneous geodesic confidence tubes with asymptotically exact coverage under mild conditions. Simulations show narrower active tubes in most reported settings, with improved small sample coverage in some three dimensional examples. Ocean-drifter observations illustrate mean flow reconstruction on the sphere; active sampling of canopy-flow measurements quantifies lateral and vertical departures of local mean streamlines.
\end{abstract}

\noindent\textbf{Keywords:} Functional central limit theorem; Integral curve; 
Riemannian manifold; 
Simultaneous confidence tube; Vector-field regression.

\doublespacing

\section{Introduction}\label{sec:intro}
Riemannian data arise when the objects being measured have geometric constraints: directions lie on spheres, orientations are rotations, and many shape representations form nonlinear spaces. Respecting this geometry is essential for comparing observations and describing their variation \citep{pennec2006intrinsic}. When the observations describe local velocities, integral curves connect these measurements to pathways and transport. Examples include direction-field trajectories in diffusion imaging and mean circulation inferred from ocean drifters \citep{koltchinskii2007integral,lumpkin2013global}. The statistical task is to recover a trajectory together with its uncertainty from noisy information. 

Wall shear stress on a curved body provides a concrete example of this problem. The shear stress is tangent to the surface, and its integral curves, called skin-friction lines, describe the organization of near wall flow \citep{simmons2022experimental}. Recovering a finite segment of such a curve from sparse, noisy vector measurements requires accounting for both surface geometry and measurement uncertainty. Accurate shear measurements are often available only at a limited number of sensor locations \citep{rohlfs2025sense}. Since the target line is unknown in advance, a sparse pilot survey can first provide an approximate route. When further measurement locations can be chosen, the remaining budget can be concentrated along and across this estimated route to refine the curve while allowing it to depart from the pilot. Adaptive acquisition is also used in laser Doppler anemometry (LDA), Doppler lidar and underwater sampling \citep{theunissen2015autonomous,wildmann2018scanning,kemna2018onboard}, while streamlines inform motion planning in ocean currents \citep{to2019streamlines}. Together with existing archives such as ocean-drifter data, these settings motivate a distinction between using available observations and directing new measurements toward a curve of interest.

We formulate these questions as a statistical problem of estimating an
integral curve from noisy vector measurements. At a recorded position
$X_i$, the tangent vector response follows
\begin{equation}\label{eq:regression-model}
 Y_i=V(X_i)+\varepsilon_i\in T_{X_i}\M,
\end{equation}
where $\M$ is a known Riemannian manifold, $V$ is an unknown vector field and the errors are conditionally centered. The target is its parameterized integral curve from a given starting point over a fixed time interval. This leads to two questions.
\begin{itemize}
    \item Q1: With independent random-design observations already available, how can we estimate the entire curve and construct a simultaneous confidence tube?
    \item Q2: If a total response budget allows us to choose new locations, how can a preliminary sample guide those measurements to improve accuracy while allowing valid simultaneous inference?
\end{itemize}

\begin{figure}[h]
\centering
\includegraphics[width=\textwidth]{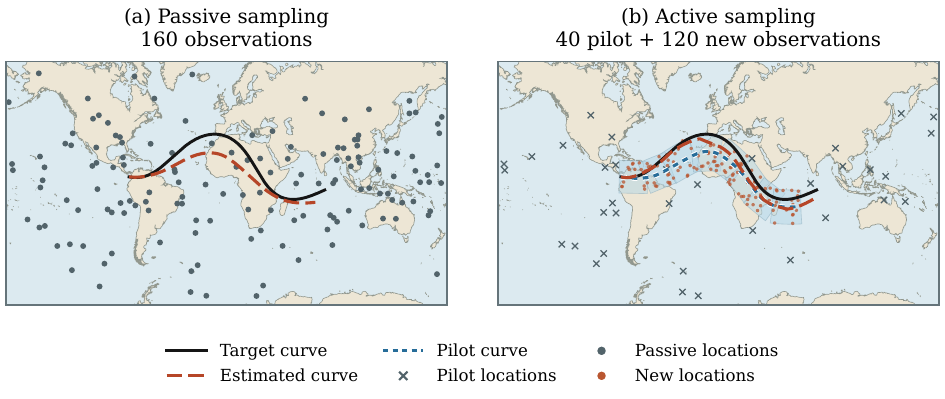}
\caption{Simulated atmospheric-flow reconstruction under passive and active sampling.}
\label{fig:sampling-policies}
\end{figure}

Recovering a mean wind trajectory from local velocity measurements identifies an average atmospheric transport route over both land and ocean. Figure~\ref{fig:sampling-policies} uses simulated velocities to illustrate passive estimation and pilot-guided sampling and reconstruction on Mercator maps ($70^\circ$S--$70^\circ$N), with Natural Earth land boundaries.

The first question has substantial Euclidean precedents. \citet{koltchinskii2007integral} study noisy vector observations at uniform random locations and derive a curve functional central limit theorem with linear stochastic error propagation. \citet{koltchinskii2009asymptotics} study curve functionals, and \citet{sakhanenko2011global} establishes global minimax rates in that regression setting{. Later work studies how to accelerate integral curve estimation \citep{sakhanenko2015acceleration}.} Tensor based models instead infer a direction field from noisy tensor measurements; their curve estimation risks are studied by \citet{banerjee2023global,banerjee2024supplement}. These observation models must be distinguished from direct tangent vector regression. Likewise, \citet{ariascastroqiao2026bands} construct confidence regions for density-gradient lines from observations of positions alone. The derivative of a density estimator, rather than an observed velocity, drives those curves. {Related sampling strategies arise in randomized initial value solvers, which select where to evaluate the field or its derivatives \citep{kacewicz2006almost}, and in scalar inverse regression, where a two-stage design yields a parametric rate and bootstrap inference \citep{tang2011two}.}

On manifolds, \citet{pelletier2005density,pelletier2006regression} provides volume corrected kernel estimators for density and scalar regression. \citet{linyao2019intrinsic} map observed trajectories to tangent vectors along their Fr\'echet mean curve to identify principal modes of variation. Furthermore, \citet{linyao2021contamination} use local linear regression on a manifold to predict a scalar response from a noisy functional predictor. In our model, we observe noisy tangent velocities at sampled locations and recover an unobserved trajectory by integrating the estimated field. Parallel transport lets us average velocities in different tangent spaces, but field errors must then be propagated through the dynamics. The general evolution equation theory of \citet{weipanaretos2018empirical} explains this transfer and provides bootstrap results when the scaled field estimation error has a weak limit. For shrinking-bandwidth tangent regression, we additionally establish the integrated error law and feasible covariance along the random fitted curve.

To answer Q1, we integrate an intrinsic kernel estimate of $V$ and prove a functional central limit theorem for the estimation error of the entire integral curve. After rescaling and parallel transport to a common tangent space, the error process converges to a Gaussian process that solves a linear stochastic differential equation. Its drift describes how the field propagates earlier errors and incorporates the smoothing bias, while its diffusion describes how velocity noise accumulates along the curve. The curve rate $n^{-2/(d+3)}$ agrees with the corresponding second-order Euclidean rate. Under undersmoothing, we use a Gaussian approximation at the fitted bandwidth and consistently estimate its covariance across times to construct simultaneous geodesic confidence tubes. In flat coordinates, transport is the identity and these constructions become normalized Euclidean kernel regression and its propagated error. 

{To answer Q2, we use an independent kernel pilot to locate the curve and collect one batch of location-response pairs in neighborhoods along the pilot.} Sampling across each neighborhood identifies transverse changes in the field, allowing the reconstructed curve to depart from the pilot. Conditional on successful localization, the final responses determine the leading estimation error. Under the stated smoothness and design conditions, the expected maximum curve error is $O(N^{-1/2})$, with pilot responses included in $N$. A kernel pilot and a fixed pilot fraction suffice in dimensions two and three at the stated smoothness.

For this active experiment, we establish simultaneous confidence tubes whose radii have order $N^{-1/2}$ and whose coverage tends to the nominal level under undersmoothing and nondegenerate response noise. Thus concentrating observations near the curve improves estimation accuracy while retaining uncertainty quantification for the entire trajectory. The inference remains valid across changing local sampling designs, without requiring their covariances to approach a single limit. The simulations compare the two procedures at common response budgets: active tubes are narrower in most settings and improve small sample coverage in some three dimensional examples. The applications examine how observational support limits a regional mean flow curve on the sphere and use pilot-guided LDA reconstruction to quantify the stronger lateral and downward departure of local mean flow within a model canopy than above it.

The rest of the paper is organized as follows. Section~\ref{sec:geometry} gives the model and geometric assumptions. Sections~\ref{sec:passive} and~\ref{sec:cloud} develop the passive and active methods with theoretical guarantees and algorithms to answer Q1 and Q2. Sections~\ref{sec:numerics} and~\ref{sec:appdesign} report simulations and the applications. Implementation details, additional experiments and supporting application analyses appear in Appendices~\ref{app:numerics}--\ref{app:appdesign}. Proofs are provided in the supplementary material.

\section{Model and geometry}\label{sec:geometry}

Let $(\M,g)$ be a fixed, connected, compact $C^\infty$ Riemannian manifold
without boundary, of dimension $d\ge2$, with distance
$\rho$, Levi-Civita connection $\nabla$, exponential map $\Exp$, and
Riemannian volume $\mathrm{vol}$.  If $x$ and $y$ lie in a common normal
neighborhood, $\cP_{y\to x}:T_y\M\to T_x\M$ denotes parallel transport along
their unique minimizing geodesic.  Transport along a specified curve is
written separately below.

Under passive sampling, the pairs in \eqref{eq:regression-model} are independent, with $X_i\sim p\,\mathrm{vol}$, $\E(\eps_i\mid X_i)=0$ and
$\E(\eps_i\otimes\eps_i\mid X_i=x)=\Sigma(x)$.
Here $p$ is the design density and $\Sigma$ is the conditional response covariance.
{The pilot sample in the active design consists of independent copies of $(X,Y)$ from the same regression model, with $X\sim p\,\mathrm{vol}$.} Throughout, conditionally centered errors are sub-Gaussian in unit tangent directions:
\[
 \E\bigl(e^{u\langle w,\varepsilon_i\rangle}\mid X_i=x\bigr)
 \le e^{\sigma^2u^2/2},\qquad \|w\|=1,\quad u\in\R.
\]
{The same finite $\sigma$ applies to all laws in the model class and need not be known.} For final active
responses, the same bound is conditional on the pilot and all recorded
positions. Fix $0<T<\infty$ and a starting point $x_0\in\M$. The target is
\begin{equation}\label{eq:target-curve}
 \dot x(t)=V(x(t)),\qquad x(0)=x_0,\qquad 0\le t\le T.
\end{equation}
Its parameterization is determined by $V$: normalizing the field would generally change the target. Let $R_t:T_{x(t)}\M\to T_{x_0}\M$ denote parallel transport along $x$ from time $t$ to time zero. Its inverse $R_t^{-1}$ transports vectors back to $T_{x(t)}\M$. We represent accumulated curve error in $T_{x_0}\M$.

Smoothness norms use a fixed finite atlas with compact coordinate domains; for noninteger orders they include the corresponding H\"older seminorm. The known geometry determines fixed bounds for coordinate maps and their derivatives.

\begin{assumption}\label{ass:geometry}
The density $p$ and field $V$ are $C^5$, and $\Sigma$ is $C^3$.
The density is bounded away from zero, and $\Sigma$ is uniformly
positive definite.
\end{assumption}

\begin{assumption}\label{ass:kernel}
The nonnegative radial function \(K:\R^d\to\R\) is \(C^4\), supported in the
open unit ball, and satisfies
$
 \int_{\R^d}K(u)\dd u=1,
 \int uK(u)\dd u=0,
 \int uu^\top K(u)\dd u=\mu_2(K)I_d.$
\end{assumption}

\begin{assumption}\label{ass:curve}
The solution $x:[0,T]\to\M$ is an embedding and
$\inf_{0\leq t\leq T}\norm{V(x_t)}>0$.
{There is a constant $c_{\rm nr}>0$, common to all laws in the model class, such that}
\begin{equation}\label{eq:bilip}
 \rho(x_s,x_t)\geq {c_{\rm nr}}\abs{s-t},\qquad s,t\in[0,T].
\end{equation}
\end{assumption}
\begin{assumption}\label{ass:pilot-model}
For some
$\nu>2$, the admissible laws have fixed known bounds $
 p\ge p_*>0,
 \|p\|_{C^\nu}\le M_p,\|V\|_{C^\nu}\le M_V.$
Conditional on the pilot and all recorded positions, the final response errors are mutually independent.
\end{assumption}

The density, smoothness and kernel assumptions are familiar from nonparametric regression on manifolds. A design density bounded away from zero gives each neighborhood positive sampling probability, and a smooth regression function controls the bias of local averages \citep{pelletier2006regression}. \citet{chengwu2013local} likewise use a compactly supported radial kernel and allow the conditional response variance to vary with location. Here radial symmetry cancels the first-order bias, compact support keeps parallel transport within normal neighborhoods, and the sub-Gaussian bound controls fluctuations of the local averages.

Recovering an integral curve also requires controlling how velocity errors accumulate along it. For passive inference, the additional smoothness in Assumption~\ref{ass:geometry} controls derivatives of the fitted field, and the covariance lower bound ensures nondegenerate uncertainty. Assumption~\ref{ass:curve} limits the time the curve spends in any kernel neighborhood, as in the Euclidean integral curve model of \citet[Theorem~1]{koltchinskii2007integral}. It excludes stationary points and returns over the observation interval. For active sampling, the known bounds in Assumption~\ref{ass:pilot-model} determine the sampling radii and fit thresholds. Active inference in Section~\ref{sec:active-inference} additionally requires a lower bound on the response covariance.

\textit{Notation.} Let $M_1$ denote the common intrinsic $C^1$ bound on $V$ and $D_\M$ the diameter of $\M$. For curves $x$ and $y$, write $d_\infty(y,x)=\sup_{0\le t\le T}\rho\bigl(y(t),x(t)\bigr)$.

\section{Passive sampling for Q1}\label{sec:passive}

We first consider observations collected at independent random locations.
Intrinsic kernel regression estimates the vector field, and integrating
the fitted field gives the curve estimate. We derive its limiting error
process and use that law to construct simultaneous confidence tubes.

\subsection{Curve estimation}
The estimator averages nearby velocity measurements after parallel transport
to a common tangent space, then integrates the fitted field from $x_0$.
The weights correct for Riemannian volume and stabilize small denominators.
In normal coordinates at
\(x\), write \((\Exp_x)^*\mathrm{vol}=\theta_x(u)\dd u\). For a radial kernel
\(K\) as in Assumption~\ref{ass:kernel} and \(h\) smaller than the injectivity radius,
fix \(0<\varpi<\inf_{x\in\M}p(x)\) and a smooth function
\(s_\varpi:[0,\infty)\to(0,\infty)\) satisfying
\(s_\varpi(r)=r\) for \(r\geq\varpi\) and
\(s_\varpi(r)\geq\varpi/2\), and define
\begin{align}
 W_h(x,y)&=
 \begin{cases}
 h^{-d}K\bigl(\Log_x y/h\bigr)/\theta_x(\Log_x y),
     &\rho(x,y)<\operatorname{inj}(\M),\\
 0,&\rho(x,y)\geq\operatorname{inj}(\M),
 \end{cases}\label{eq:weight}\\
 \wh p_h(x)&=\frac1n\sum_{i=1}^nW_h(x,X_i),\qquad
 \wh N_h(x)=\frac1n\sum_{i=1}^nW_h(x,X_i)\cP_{X_i\to x}Y_i,\label{eq:pn}\\
 \wh V_h(x)&=\frac{\wh N_h(x)}{s_\varpi\bigl(\wh p_h(x)\bigr)}.
 \label{eq:estimator}
\end{align}
The replacement of $\wh p_h$ by $s_\varpi(\wh p_h)$ is a smooth
stabilization: it prevents division by a density estimate close to zero,
while leaving the ratio unchanged wherever $\wh p_h\ge\varpi$.
Uniform consistency of $\wh p_h$ and $\inf p>\varpi$ make this
replacement inactive with probability tending to one. Volume correction makes the
change of variables $y=\Exp_x(hu)$ exact and removes curvature from the
second-order scalar integration formula; it is standard in intrinsic kernel
estimation \citep{pelletier2005density,pelletier2006regression}.

The plug-in estimate of \eqref{eq:target-curve} solves
\[
 \dot{\widehat x}(t)=\widehat V_h(\widehat x(t)),\qquad
 \widehat x(0)=x_0,\qquad 0\le t\le T.
\]

For the passive estimator we use Assumptions~\ref{ass:geometry}--\ref{ass:curve}
and the following bandwidth condition.

\begin{assumption}\label{ass:bandwidth}
As $n\to\infty$,
\begin{equation}\label{eq:bandwidth}
 h\to0,\qquad
 \frac{(\log n)^5}{nh^{d+2}}\to0,
 \qquad nh^{d+3}\to\beta\in[0,\infty).
\end{equation}
\end{assumption}
The first two conditions ensure that both $\widehat V_h$ and its
spatial derivative $\nabla\widehat V_h$ converge uniformly to $V$ and
$\nabla V$, respectively. The derivative
describes how velocity changes with position, and hence how an error in
the curve affects its subsequent motion. The last condition controls
the accumulated smoothing bias; confidence tubes use $\beta=0$.
The same balance $nh^{d+3}\to\beta$ appears in
\citet[Theorem~1]{koltchinskii2007integral}, together with
$nh^{d+2}\to\infty$. Our logarithmic strengthening supports uniform
control of the fitted field and its derivative.
{The exponent five in \eqref{eq:bandwidth} suffices to
control the derivatives of the kernel numerator and denominator, the
interpolation error within coordinate charts, and the remainder along
the trajectory.}

\subsection{Curve error and simultaneous inference}\label{sec:integrated}

We first identify the distribution of the error accumulated along the
estimated curve. After scaling and parallel transport to a common
tangent space, this error converges to a Gaussian process satisfying a
linear stochastic differential equation. This functional limit provides
the basis for the simultaneous inference developed below.

Put $a_n=(nh^{d-1})^{1/2}$ and $A_t=R_t(\nabla V)_{x_t}R_t^{-1}$.
Let $\Phi(t,s)$ solve $\partial_t\Phi(t,s)=A_t\Phi(t,s)$ with $\Phi(s,s)=I$;
it propagates an error made at time $s$ to time $t$.
When the estimated curve stays in a common normal tube around $x$,
define $Z_n(t)=a_nR_t\Log_{x_t}\wh x_t$. The logarithm represents the
geodesic displacement from the true curve, and $R_t$ puts these errors
in one tangent space so that they can be compared across time. Outside
this tube the short logarithm need not be uniquely defined, so we set
$Z_n=0$ to define the process for every sample. This exceptional event has probability
tending to zero and does not affect the limiting distribution.

Define the connection-Laplacian bias field
\begin{equation}\label{eq:bias}
 B(x)=\mu_2(K)\left(\frac12\Delta^\nabla V(x)
       +\frac1{p(x)}\nabla_{\operatorname{grad}p(x)}V\right).
\end{equation}
The sign convention is $\Delta^\nabla=\tr_g\nabla^2$, where $\tr_g\nabla^2$ is the trace of the second covariant derivative with respect to the Riemannian metric $g$.

For $a\ne0$, put
\begin{equation}\label{eq:psi}
 \psi_K(a)=\int_{\R}\int_{\R^d}K(z)K(z-\tau a)\dd z\dd\tau.
\end{equation}
The integral is invariant under orthonormal changes of frame and therefore
also defines $\psi_K$ for nonzero tangent vectors.

\begin{theorem}[Integral-curve functional central limit theorem]
\label{thm:curve}
Under Assumptions~\ref{ass:geometry}--\ref{ass:curve} and~\ref{ass:bandwidth},
$Z_n\Rightarrow Z$ in $C([0,T],T_{x_0}\M)$, where, for standard Brownian
motion $W$ in $T_{x_0}\M$, $Z_0=0$ and
\begin{equation}\label{eq:limitSDE}
 \dd Z_t=A_tZ_t\dd t+\sqrt\beta\,R_tB(x_t)\dd t+\left[\frac{\psi_K(V(x_t))}{p(x_t)}R_t\Sigma(x_t)R_t^{-1}\right]^{1/2}\dd W_t.
\end{equation}
Equivalently, with the transition matrix $\Phi(t,s)$ defined above,
\begin{equation}\label{eq:variation}
 Z_t=\sqrt\beta\int_0^t\Phi(t,s)R_sB(x_s)\dd s
       +\int_0^t\Phi(t,s)\left[\frac{\psi_K(V(x_s))}{p(x_s)}R_s\Sigma(x_s)R_s^{-1}\right]^{1/2}\dd W_s.
\end{equation}
Its mean $m_t$ and covariance $C_t$ are the unique solutions of
\begin{align}
 \dot m_t&=A_tm_t+\sqrt\beta\,R_tB(x_t),&m_0&=0,\label{eq:meanode}\\
 \dot C_t&=A_tC_t+C_tA_t^*+\frac{\psi_K(V(x_t))}{p(x_t)}R_t\Sigma(x_t)R_t^{-1},&C_0&=0.\label{eq:covode}
\end{align}
For $s\leq t$,
$\Cov(Z_s,Z_t)=C_s\Phi(t,s)^*$.
\end{theorem}

The drift $A_tZ_t$ propagates earlier errors through the dynamics,
while the diffusion term adds the local contribution of response noise.
In a flat chart, $R_t=I$, $\theta_x=1$ and $\nabla$ is the ordinary
derivative. Unlike the known-density numerator estimator in
\citet{koltchinskii2007integral}, the normalized ratio here removes the
leading design fluctuation proportional to $VV^*$; its diffusion
coefficient therefore involves the response covariance $\Sigma$.
The theorem also includes the leading smoothing bias when $\beta>0$.
For confidence tubes we undersmooth so that $\beta=0$ and the limiting
process is centered. Since $\|Z_n\|_\infty=a_n d_\infty(\widehat x,x)$
on the normal tube, a quantile of $\|Z\|_\infty$, divided by $a_n$,
describes the radius needed to cover the entire curve. This connects the curve error to the radius of a simultaneous tube.

We now choose the bandwidth for the second-order estimator of
Section~\ref{sec:passive}. Its curve error combines a bias of order
\(h^2\) and a random term of order \((nh^{d-1})^{-1/2}\).
The next result gives the balanced estimation rate. All probability statements in this section are for a
fixed model satisfying the passive assumptions.

\begin{corollary}\label{cor:passive-rates}
Under Assumptions~\ref{ass:geometry}--\ref{ass:curve}, let \(d>1\).
If \(h\asymp n^{-1/(d+3)}\), then
\[
 \sup_{t\le T}\rho(\wh x_t,x_t)=O_{\PP}(n^{-2/(d+3)}).
\]

\end{corollary}

Balancing $h^2$ and $(nh^{d-1})^{-1/2}$ gives
$h\asymp n^{-1/(d+3)}$. This minimizes the upper bound on curve error;
the error can be smaller when the leading bias vanishes.
For centered inference, choose a smaller bandwidth satisfying both
$nh^{d+3}\to0$ and $(\log n)^5/(nh^{d+2})\to0$.
The latter condition prevents undersmoothing so far that derivative noise
invalidates the curve linearization. The logarithmic power is sufficient for our uniform bounds.

The curve rate $n^{-2/(d+3)}$ is faster than the pointwise field rate
$n^{-2/(d+4)}$ for second-order kernel regression
\citep{stone1982optimal}. Thus estimating the trajectory can be more
accurate than estimating the field at an individual location. The curve
rate agrees with the Euclidean result of \citet{sakhanenko2011global},
which establishes minimax bounds under different regression assumptions.
Corollary~\ref{cor:passive-rates} gives a probability bound for each fixed
manifold model.

The limiting SDE explains which error paths must be reproduced, but its
coefficients depend on the unknown field, density and response covariance.
We construct a computable Gaussian approximation by propagating
observation-level perturbations through the fitted dynamics. At the
chosen bandwidth, one observation can affect several nearby points on
the curve. Using the same perturbation throughout the trajectory preserves
this dependence. Under undersmoothing, the resulting scaled process
approximates the curve error and has the same limit as the SDE in
Theorem~\ref{thm:curve}.

We first define the Gaussian law whose supremum quantile will determine
the tube radius. Its covariance includes each observation's contribution
at bandwidth $h$.
Put $p_h^0(x)=\E W_h(x,X)$ and
$\bar V_h(x)=\E\bigl(W_h(x,X)\cP_{X\to x}Y\bigr)/p_h^0(x)$.
For a possible observation $(z,y)$ define its propagated contribution
\begin{equation}\label{eq:passive-influence}
 J_h(t;z,y)=\int_0^t\Phi(t,s)R_s
 \frac{W_h(x_s,z)}{p_h^0(x_s)}
 \bigl(\cP_{z\to x_s}y-\bar V_h(x_s)\bigr)\dd s.
\end{equation}
It has mean zero. Let $\mathbb G_h$ be the centered continuous Gaussian
process on $T_{x_0}\M$ with covariance
\begin{equation}\label{eq:finite-covariance}
 C_h(s,t)=h^{d-1}\E\bigl(J_h(s;X,Y)\otimes J_h(t;X,Y)\bigr).
\end{equation}
Both the random positions and the conditional response noise contribute
to this covariance. The kernel overlap at two different times is kept
inside the expectation.
The bounded-Lipschitz distance $d_{\rm BL}$ below uses the supremum norm
on $C([0,T],T_{x_0}\M)$.

\begin{theorem}\label{thm:passive-finite}
Under Assumptions~\ref{ass:geometry}--\ref{ass:curve} and~\ref{ass:bandwidth} and $
 nh^{d+3}\longrightarrow0,
$
the following uniform expansion and Gaussian approximation hold:
\begin{align}
 \left\|Z_n-\frac{a_n}{n}\sum_{i=1}^n
 J_h(\,\cdot\,;X_i,Y_i)\right\|_\infty&=o_{\PP}(1),
 \label{eq:finite-linearization}\\
 d_{\rm BL}\bigl(\mathcal L(Z_n),\mathcal L(\mathbb G_h)\bigr)&\longrightarrow0.
 \label{eq:finite-gaussian}
\end{align}
Moreover, $
 \sup_{r\in\R}\left|
 \PP(\|Z_n\|_\infty\le r)-
 \PP(\|\mathbb G_h\|_\infty\le r)\right|\longrightarrow0.$
\end{theorem}

Theorem~\ref{thm:passive-finite} approximates the distribution of the entire
path and its maximum. The Gaussian law depends on $h$, and the
approximation is asymptotic as $n$ grows.
Its limiting covariance is that of the centered SDE in
\eqref{eq:limitSDE}.

To approximate the curve error, we perturb the observation weights and
responses by independent Gaussian variables and propagate these
perturbations through the fitted dynamics. Each observation uses the
same perturbation throughout time, preserving the dependence between
overlapping kernel neighborhoods. This reproduces entire error paths,
so their maximum displacement can determine a simultaneous tube radius.

We estimate the auxiliary mean and response covariance without using
an observation in its own fit. Choose a fixed number of outer folds
$F_1,\ldots,F_L$, independently of the observations, and partition each
complement $\mathcal I_\ell=\{1,\ldots,n\}\setminus F_\ell$ into inner
validation sets $J_{\ell f}$. All training and validation sizes are at
least $cn$ for a fixed $c>0$. Let $\wh V_{I,b}$ denote
\eqref{eq:estimator} fitted to sample $I$, with bandwidth $b$ and
normalization $|I|$. For $i\in F_\ell$ and $j\in J_{\ell f}$, put
\begin{equation}\label{eq:arxiv-auxiliary}
 M_i=\wh V_{\mathcal I_\ell,b}(X_i),\qquad
 e_{\ell j}=Y_j-\wh V_{\mathcal I_\ell\setminus J_{\ell f},b}(X_j).
\end{equation}
Pool the inner residuals to estimate their covariance:
\begin{equation}\label{eq:nested-covariance}
 S_\ell^{\rm raw}(x)=
 \frac{|\mathcal I_\ell|^{-1}\sum_{j\in\mathcal I_\ell}W_g(x,X_j)
 (\cP_{X_j\to x}e_{\ell j})(\cP_{X_j\to x}e_{\ell j})^*}
 {s_\varpi\bigl(|\mathcal I_\ell|^{-1}\sum_{j\in\mathcal I_\ell}W_g(x,X_j)\bigr)}.
\end{equation}
For $i\in F_\ell$, let $S_i$ be $S_\ell^{\rm raw}(X_i)$ with its
eigenvalues truncated below at $\gamma_n\downarrow0$. The auxiliary
bandwidths satisfy
\begin{equation}\label{eq:pilotbw}
 b,g\to0,\qquad \frac{\log n}{nb^d}\to0,\qquad
 \frac{\log n}{ng^d}\to0.
\end{equation}

The perturbation weights account for the stabilized denominator. Put
\[
 \alpha_i(x)=\frac{W_h(x,X_i)}{n s_\varpi(\wh p_h(x))},\qquad
 \bar M(x)=\sum_i\alpha_i(x)\cP_{X_i\to x}M_i,
\]
\[
 \beta_i(x)=s_\varpi'(\wh p_h(x))\alpha_i(x)
 +\frac{1-s_\varpi'(\wh p_h(x))\wh p_h(x)/s_\varpi(\wh p_h(x))}{n},
\]
and define the position and response perturbations by
\[
 D_i^{\rm pos}(x)=\alpha_i(x)\cP_{X_i\to x}M_i-\beta_i(x)\bar M(x),
 \qquad N_i(x)=\alpha_i(x)\cP_{X_i\to x}S_i^{1/2}.
\]
Here $\sum_iD_i^{\rm pos}(x)=0$, so the position perturbations are centered.
Let $\wh R_t$ transport along $\wh x$ back to $T_{x_0}\M$, and set
$\wh A_t=\wh R_t(\nabla\wh V_h)_{\wh x_t}\wh R_t^{-1}$.
Draw independent $\zeta_i\sim N(0,1)$ and standard Gaussian vectors
$\eta_i\in T_{X_i}\M$, independently of the data. Keeping these draws
fixed across time, solve
\begin{equation}\label{eq:finite-bootstrap}
 \dot Z_n^\#(t)=\wh A_tZ_n^\#(t)+
 \sum_{i=1}^n\wh R_t\bigl(D_i^{\rm pos}(\wh x_t)\zeta_i
                           +N_i(\wh x_t)\eta_i\bigr),\quad
 Z_n^\#(0)=0.
\end{equation}
This ordinary differential equation propagates both sources of variation
through the fitted curve dynamics. Its solution is in unscaled
curve-error units; write
$\wh C_n(s,t)=\Cov^\#(Z_n^\#(s),Z_n^\#(t))$ for its covariance
conditional on the data.
Let $R_{n,1-\alpha}$ be the conditional $(1-\alpha)$-quantile of
$\|Z_n^\#\|_\infty$. The simultaneous tube is
$\mathcal T_n(t)=\{y:\rho(y,\wh x_t)\le R_{n,1-\alpha}\}$.

The following theorem connects the simulation to the curve error:
the scaled covariance $a_n^2\wh C_n$ consistently approximates $C_h$, and the
conditional law of $a_nZ_n^\#$ approximates the same Gaussian law as
$Z_n$. The supremum quantile therefore gives simultaneous coverage.
Estimating only $\wh C_n(t,t)$ would miss the cross-time dependence
that determines a maximum over the trajectory. The supremum over pairs
$(s,t)$ below controls that dependence even though propagation is evaluated
on the same random curve that is being estimated.

\begin{theorem}\label{thm:inference}
Under the conditions of Theorem~\ref{thm:passive-finite} and
\eqref{eq:pilotbw}, the preceding estimators satisfy
\begin{align}
 \sup_{s,t\le T}\left\|a_n^2\wh C_n(s,t)-C_h(s,t)\right\|&\to_{\PP}0,
 \label{eq:finite-cov-consistency}\\
 d_{\rm BL}\bigl(\mathcal L^\#(a_nZ_n^\#),\mathcal L(\mathbb G_h)\bigr)
 &\to_{\PP}0.
 \label{eq:finite-bootstrap-validity}
\end{align}
The conditional distribution function of $a_n\|Z_n^\#\|_\infty$
approximates that of $\|Z_n\|_\infty$ uniformly over thresholds,
in probability.
Consequently,
\begin{equation}\label{eq:tubecoverage}
 \PP\{x_t\in\mathcal T_n(t)\text{ for all }t\le T\}\to1-\alpha.
\end{equation}
\end{theorem}

Thus the CLT identifies the error dynamics, and the finite-bandwidth
Gaussian approximation allows quantiles of the maximum curve error to be estimated. Taking the
maximum of each simulated error path accounts for uncertainty over the
 whole time interval and gives one radius for the simultaneous tube.

\begin{algorithm}[H]
\caption{Passive curve estimation and simultaneous tube}\label{alg:passive}
\begin{algorithmic}[1]
\Require Data, $x_0,T$, stabilization $s_\varpi$, level $1-\alpha$, and
bandwidths satisfying \eqref{eq:bandwidth} with $\beta=0$ and \eqref{eq:pilotbw}.
\State Fit \eqref{eq:estimator} and integrate from $x_0$ to obtain $\wh x$.
\State Compute auxiliary means and covariances using \eqref{eq:arxiv-auxiliary}--\eqref{eq:nested-covariance}.
\State Simulate \eqref{eq:finite-bootstrap}, keeping each Gaussian draw fixed across time;
record $\|Z_n^\#\|_\infty$.
\State Take its $(1-\alpha)$-quantile $R_{n,1-\alpha}$; return $\wh x$ and
$\{y:\rho(y,\wh x_t)\le R_{n,1-\alpha}\}$.
\end{algorithmic}
\end{algorithm}

A numerical implementation approximates the supremum on an evaluation
grid and the quantile with finitely many Gaussian draws.

\section{Active sampling for Q2}\label{sec:cloud}

When measurement locations can be controlled, new responses can be
concentrated near the route of interest. The route is unknown, so an
initial passive sample first provides a pilot curve. We then use that
curve to choose sampling neighborhoods and distribute the remaining
measurements throughout them. The neighborhoods allow the final curve
to depart from the pilot while keeping the new observations relevant
to its reconstruction. The total response budget is $N$: an independent
pilot uses $n_0$ observations, and the final batch uses $m=N-n_0$ new
responses. All acquired responses count toward the budget, including
those in rejected fits.

\subsection{The design}\label{sec:active-strategy}

The true route is unknown, so the pilot indicates where new measurements
will be informative. Its error must be small relative to the sampling
neighborhoods; otherwise those measurements may miss the true route.
We require the pilot to locate the true curve to within $o(\Delta)$,
where $\Delta=T/J$ is the time-segment length and the reconstruction
neighborhoods have radius proportional to $\Delta$.
Fit \eqref{eq:estimator} to the first $n_0$ observations with bandwidth $b=b_{n_0}$, and let $\gamma_0$ be its integral curve from $x_0$. Use a kernel as in Assumption~\ref{ass:kernel} that is positive near zero, take $b$ below the common normal radius, and fix $0<\varpi<p_*$.

For $b\asymp n_0^{-\beta}$ with fixed $0<\beta<1/d$, put
\begin{equation*}
 \epsilon_0=b^2+\sqrt{\frac{\log(2n_0)}{n_0b^d}},\qquad
 H_0=n_0b^d.
\end{equation*}
Every fixed moment of $d_\infty(\gamma_0,x)$ is bounded by the corresponding
power of $\epsilon_0$, plus an exponentially small term in $H_0$.
Thus the budget and bandwidth can make the pilot accurate relative to
$\Delta$; the final fit need not reproduce the pilot's estimation error.

The pilot determines the sampling neighborhoods and the allocation of
the remaining observations. Within these neighborhoods, we collect new
location-response pairs under the random sampling strategy below.

Dividing the time interval into short segments lets us reconstruct each
part of the route from measurements in a small neighborhood. Allocating
observations to every segment supplies information along the whole curve.
Let $J$ be a positive integer, $\Delta=T/J$, $t_\ell=\ell \Delta$, and
$I_\ell=[t_\ell,t_{\ell+1})$, closing the last cell at $T$.
For each cell, center a sampling neighborhood at $\gamma_0(t_\ell)$
and use orthonormal coordinates on $T_{\gamma_0(t_\ell)}\M$.
We acquire responses at positions of the form
$X_i=\Exp_{\gamma_0(t_\ell)}(\Delta z_i)$
with bounded scaled coordinates $|z_i|\le K$, where $K$ is fixed.
Thus the pilot specifies the regions to which a mobile sensor is sent
or a scanning instrument is directed. For sufficiently small $\Delta$,
all these positions lie inside their designated charts by construction.
The pilot determines the coordinate charts and the conditional sampling
laws within them.

Choose a fixed reconstruction radius $R$ in scaled coordinates large
enough to cover the movement during one segment and the pilot's location
error. {On $A_\Delta=\{d_\infty(\gamma_0,x)\le c_{\rm nr}\Delta/16\}$, each true segment lies inside its reconstruction neighborhood.}
Allocate the remaining $m=N-n_0$ responses so that the cell counts
$n_\ell=\#\{i:\ell_i=\ell\}$ satisfy $\sum_\ell n_\ell=m$ and
$n_\ell\asymp m/J\asymp m\Delta$ uniformly over cells, with comparison
constants independent of $N$ and $\ell$. This allows nonuniform
allocation, including proportions chosen from the pilot, while giving
each segment a sample size of the same order. Each observation retains
its acquisition cell label even when geometric neighborhoods overlap.

Within each cell, sample positions independently along the pilot chord
and in the directions transverse to it. Measurements only on the pilot
can leave transverse field changes undetermined. Spreading observations
throughout the neighborhood identifies these changes, allowing the
reconstructed curve to depart from the pilot. We fit a low-degree
polynomial to the local coordinate velocities, map the fitted velocities
back to the manifold, and integrate them to reconstruct the curve.

Write $\mathcal D$ for the pilot, final cell labels and recorded positions.
The observed pairs satisfy $Y_i=V(X_i)+\varepsilon_i$.
Conditional on $\mathcal D$, the errors
are independent, centered and uniformly sub-Gaussian. In the coordinates
of the assigned cell, the observation has position and velocity
\[
 z_i=\Delta^{-1}\Log_{\gamma_0(t_{\ell_i})}(X_i),\qquad
 Z_i=(d\Exp_{\gamma_0(t_{\ell_i}),\Delta z_i})^{-1}Y_i.
\]
Here $d\Exp_{a,u}$ denotes the differential of $\Exp_a$ at $u$.
This transformation expresses each response as the rate of change of
its local coordinates, so the regression estimates the velocity needed
to reconstruct a trajectory in that chart. With
$k=\lceil\nu\rceil-1$, fit a degree-$k$ polynomial by least squares:
\begin{equation}\label{cl:eq:fit}
 (\wh B_{\ell,\alpha})_{|\alpha|\le k}
 =\underset{(B_\alpha)_{|\alpha|\le k}}{\operatorname{argmin}}
 \sum_{i:\ell_i=\ell}
 \left|Z_i-\sum_{|\alpha|\le k}B_\alpha z_i^\alpha\right|^2.
\end{equation}
Here $\alpha$ is a multi-index and each coefficient $B_\alpha$ is a
vector in $\R^d$.

Each segment needs comparable sample counts, bounded scaled positions
and a well-conditioned polynomial Gram matrix. These ensure that the
local approximation is accurate and that the fit distinguishes its
polynomial terms without amplifying noise.
Let $\wh G_\ell$ have entries
$(\wh G_\ell)_{\alpha\beta}
=n_\ell^{-1}\sum_{i:\ell_i=\ell}z_i^\alpha z_i^\beta$ for
$|\alpha|,|\beta|\le k$. For fixed positive constants
$c_-,c_+,g_-,g_+$, the required properties are
\begin{equation}\label{cl:eq:design}
 c_-m\Delta\le n_\ell\le c_+m\Delta,\qquad |z_i|\le K,
 \qquad g_-I\preceq\wh G_\ell\preceq g_+I
\end{equation}
for every cell, with all positions in their designated charts.
Denote the event in \eqref{cl:eq:design} by $E_D$. The Gram lower bound
ensures that \eqref{cl:eq:fit} has a unique solution and controls how
observation noise affects the fitted polynomial.

Map the fitted velocities back to the manifold to obtain a local field
$\wh V_\ell$. Within the reconstruction neighborhood $|u|\le R\Delta$,
\[
 \wh V_\ell(\Exp_{\gamma_0(t_\ell)}(u))
 =d\Exp_{\gamma_0(t_\ell),u}\sum_{|\alpha|\le k}
       \wh B_{\ell,\alpha}(u/\Delta)^\alpha.
\]
Starting from $x_0$, integrate the first field and use its endpoint to
initialize the next segment:
\begin{equation}\label{cl:eq:ode}
 \dot{\wh x}(t)=\wh V_\ell(\wh x(t)),\qquad t\in I_\ell,
 \qquad \wh x(0)=x_0.
\end{equation}
The field is evaluated at the current estimated position, allowing the
reconstructed curve to move away from the pilot.

To continue the flow outside the reconstruction neighborhood, choose
a smooth function $\vartheta$ equal to one on $[0,R]$ and zero beyond
$2R$, and put $S_\Delta(u)=u\vartheta(|u|/\Delta)$.
Let $\chi$ be a smooth cutoff supported inside the chart and equal to
one on $|u|\le2R\Delta$. Extend the fitted field by
\begin{equation}\label{cl:eq:extended-field}
 \wh V_\ell(\Exp_{\gamma_0(t_\ell)}(u))
 =\chi(u)d\Exp_{\gamma_0(t_\ell),u}
   \sum_{|\alpha|\le k}\wh B_{\ell,\alpha}
                      (S_\Delta(u)/\Delta)^\alpha,
\end{equation}
and set it to zero outside the chart. This agrees with the original fit
on $|u|\le R\Delta$ and gives a field defined along the entire flow.

We also check the fitted coefficients to control the field's magnitude
and first derivative. For a coefficient matrix $B$, define
\[
 J_R(B)=\sum_{0<|\alpha|\le k}|\alpha|(2R)^{|\alpha|-1}|B_\alpha|.
\]
The fixed model and design bounds give constants $A_0^{\rm cl}$ and
$A_1^{\rm cl}$ bounding the constant coefficient and
$\Delta^{-1}J_R$ of the conditional mean fit whenever the pilot localizes
the true curve and the design satisfies \eqref{cl:eq:design}.
Accept the response fits when, for every cell,
\begin{equation}\label{cl:eq:coefficient-checks}
 |\wh B_{\ell,0}|\le A_0^{\rm cl}+2,\qquad
 J_R(\wh B_\ell)\le(A_1^{\rm cl}+2)\Delta.
\end{equation}
These bounds keep every accepted extended field uniformly bounded
together with its first derivative. The factor $\Delta$ compensates
for the derivative of the scaled coordinate $u/\Delta$.

{Choose the count thresholds in \eqref{cl:eq:design} to contain the allocated counts. Fix $g_-$ and $g_+$ so that whenever $A_\Delta$ holds, the eigenvalues of each conditional population Gram matrix lie inside $[g_-,g_+]$ by a common positive margin. These bounds hold for sampling uniformly along and across each pilot chord.}
Algorithm~\ref{alg:active} summarizes the procedure.

\subsection{Convergence rate}\label{sec:active-risk}

The next theorem bounds the expected maximum distance between the estimated and true curves over $[0,T]$. We use this bound to choose the pilot bandwidth and segment length so that the error is $O(N^{-1/2})$, where $N=n_0+m$ counts observations from both stages.

\begin{theorem}\label{cl:thm:risk}
{Use the random neighborhood design and reconstruction of Section~\ref{sec:active-strategy} with recorded sampling positions. Under Assumptions~\ref{ass:curve} and~\ref{ass:pilot-model},}
for every fixed $p\ge2$, the neighborhood estimator satisfies, for sufficiently
small admissible $\Delta=T/J$ and $m\Delta\ge1$,
\begin{equation}\label{cl:eq:risk}
 \sup_P\E_P d_\infty(\wh x,x)
 \le C(m^{-1/2}+\Delta^\nu)
 +C_p \Delta^{-p}\bigl(\epsilon_0^p+e^{-aH_0}\bigr)
 +{C\Delta^{-1}\bigl(e^{-cm\Delta}+e^{-cm\Delta^3}\bigr)}.
\end{equation}
The constants depend on the fixed geometry, field class, noise bound,
reconstruction radius and Gram bounds.
\end{theorem}

\begin{remark}\label{rem:random-design-failure}
For the random neighborhood sampling in Section~\ref{sec:active-strategy}, the design failure probability satisfies
\[
 \sup_P\PP_P(A_\Delta\cap E_D^c)\le C\Delta^{-1}e^{-cm\Delta}
\]
under the stated count and Gram bounds. This gives the first exponential term in the last part of \eqref{cl:eq:risk}.
\end{remark}

The two-stage estimator, using a kernel pilot with $n_0\asymp N$, $m\asymp N$,
$b\asymp n_0^{-1/6}$ and $\Delta\asymp N^{-1/5}$ attains expected risk
$O(N^{-1/2})$ in dimensions two and three for $\nu=3$, under the random
sampling strategy above.
In higher dimensions, the same nonnegative kernel can be used with a
fixed pilot fraction and $b\asymp n_0^{-1/(d+4)}$. Choose
$\Delta\asymp N^{-\kappa}$ with
\[
 \frac{1}{2\nu}<\kappa<\min\left(\frac13,\frac{2}{d+4}\right).
\]
This choice yields expected risk $O(N^{-1/2})$ whenever
$\nu>\max(2,(d+4)/4)$. The sufficient smoothness threshold grows with
dimension, with degree $k=\lceil\nu\rceil-1$ used in the final local
fits; the pilot remains a second-order kernel estimate.
The allocation calculation is given in the supplementary material.

\subsection{Simultaneous confidence tubes}\label{sec:active-inference}

Section~\ref{sec:active-risk} bounds the expected maximum distance between the estimated and true curves. A tube around the estimated curve covers the true position at every time in $[0,T]$ precisely when this maximum is at most its radius. We therefore estimate the $(1-\alpha)$ quantile of the maximum distance to choose the radius.

An error in the velocity fitted on one segment can shift the starting position of the next, so errors at different times are dependent. We use the residuals from the second stage velocity fits to simulate this propagation. In each draw, we give each residual an independent Gaussian weight and follow its effect along the fitted curve, keeping the weight fixed over time. We record the largest displacement along each simulated path and use the resulting quantile as the tube radius.

Order the polynomial coefficients with the constant term first and
write $\wh\theta_\ell=\operatorname{vec}(\wh B_\ell)$, using columnwise
vectorization. Set
\[
 D_i=(z_i^\alpha I_d)_{|\alpha|\le k},\qquad
 Q_\ell=\sum_{i:\ell_i=\ell}D_i^\top D_i,\qquad
 r_i=Z_i-D_i\wh\theta_{\ell_i}.
\]
The influence matrix $\wh\Psi_i(t)$ maps residual $i$ to its propagated
displacement at time $t$. To compute it, let $\wh R_t$ transport back
along $\wh x$, identify $T_{x_0}\M$ with $\R^d$ using a fixed
orthonormal basis, and solve
\[
 \wh A(t)=\wh R_t(\nabla\wh V_{\ell(t)})_{\wh x(t)}\wh R_t^{-1},\qquad
 \partial_t\wh\Phi(t,s)=\wh A(t)\wh\Phi(t,s),\quad\wh\Phi(s,s)=I.
\]
For $u\in I_\ell$, put $\wh w=\Log_{\gamma_0(t_\ell)}(\wh x(u))$.
The coefficient derivative of \eqref{cl:eq:extended-field} is
\[
 \wh T_\ell(u)[\operatorname{vec}(B)]
 =\wh R_u\chi(\wh w)d\Exp_{\gamma_0(t_\ell),\wh w}
   \sum_{|\alpha|\le k}B_\alpha(S_\Delta(\wh w)/\Delta)^\alpha,
\]
and is zero outside the chart. Thus
\begin{equation}\label{cl:eq:fitted-influence}
 \wh\Psi_i(t)=\left(\int_{I_{\ell_i}\cap[0,t]}
       \wh\Phi(t,u)\wh T_{\ell_i}(u)\dd u\right)
       Q_{\ell_i}^{-1}D_i^\top.
\end{equation}
This combines the change in the local least-squares fit with its
propagation through the fitted curve dynamics.
With $q=\binom{d+k}{k}$ coefficients per response component,
require $n_\ell>q$ in each accepted cell and use the degrees-of-freedom
correction, for $e_i^\#\stackrel{\rm iid}{\sim}N(0,1)$
\begin{equation}\label{cl:inf:feasible}
 \begin{split}
 \wh C_D(s,t)&=\sum_{i=1}^m\frac{n_{\ell_i}}{n_{\ell_i}-q}
       \wh\Psi_i(s)r_ir_i^\top\wh\Psi_i(t)^\top,\quad
 \widehat{\mathbb G}_N^\#(t)=\sqrt m\sum_{i=1}^m
       \sqrt{\frac{n_{\ell_i}}{n_{\ell_i}-q}}\,
       \wh\Psi_i(t)r_i e_i^\#.
 \end{split}
\end{equation}
The correction accounts for the number of fitted coefficients in each cell
without changing the fitted field or curve.
There is one scalar multiplier per vector response, shared at all times;
its law is conditional on all observed data. For $0<\alpha<1$, let
$\wh c_{1-\alpha}$ be the exact conditional $(1-\alpha)$ quantile of
$\|\widehat{\mathbb G}_N^\#\|_\infty$. Report the continuous-time tube
\begin{equation}\label{cl:inf:tube}
 \mathcal T_N(t)=\left\{y\in\M:\rho(y,\wh x(t))\le\wh c_{1-\alpha}/\sqrt m\right\}.
\end{equation}
If a chart, design or response fit is rejected, report no tube and count the attempt as noncoverage.

To construct confidence tubes, we require the local approximation bias
to be negligible relative to $m^{-1/2}$, and the noise variance to remain
bounded away from zero in every tangent direction. The following
conditions also specify the required pilot accuracy and sample sizes.

\begin{assumption}\label{cl:ass:inference}
For a fixed $\omega_->0$, every final error satisfies
$
 \Cov(\eps_i\mid\mathcal D)\succeq \omega_- I_{T_{X_i}\M}.$
As $N\to\infty$, assume $n_0,m\to\infty$, $\Delta=T/J\to0$, and
\begin{equation}\label{cl:inf:rates}
 \frac{\epsilon_0}\Delta\to0,\qquad
 \sqrt m\,\Delta^\nu\to0,\qquad
 \frac{\log(2m)}{m\Delta^3}\to0.
\end{equation}
\end{assumption}

The condition $\sqrt m\,\Delta^\nu\to0$ makes approximation bias
negligible on the curve-error scale, following the undersmoothing
principle for confidence bands in \citet[Corollary~3.1]{cck2014bands}.
The covariance lower bound provides nondegeneracy for Gaussian
supremum calibration, whose positive-variance requirement is explicit
in \citet[Theorem~3]{cck2015}. The remaining rates control pilot
localization and local-fit stability in this sampling design.

To describe the approximation, replace the fitted field and curve in
the influence matrices by the true field and curve, giving $\Psi_i$.
Let $\eps_i^c$ be the response error in the local coordinates and write
$\Omega_i=\Cov(\eps_i^c\mid\mathcal D)$. The leading error and its
conditional covariance are
$L_N(t)=\sum_i\Psi_i(t)\eps_i^c$ and
$C_D(s,t)=\sum_i\Psi_i(s)\Omega_i\Psi_i(t)^\top$.
The covariance $mC_D$ determines the possible Gaussian limits of the
normalized error. The following result states these limits and the coverage
of the residual-based tube. For formal probability statements, give the
theoretical comparison objects zero values on $\mathcal D\notin A_\Delta\cap E_D$.
Set $v_N\equiv0$ if the short logarithm between the true and estimated
curves is not available throughout $[0,T]$. On an observed chart,
design or response-fit failure, give undefined feasible covariances,
multiplier processes and critical values zero values and report no tube.
Conditioning on $\mathcal D$ always precedes response-fit acceptance.

\begin{theorem}[Active Gaussian limits and simultaneous inference]\label{cl:thm:active-inference}
For the reconstruction in Section~\ref{sec:active-strategy}, under
Assumptions~{\ref{ass:curve}, }\ref{ass:pilot-model} and~\ref{cl:ass:inference}, put
$v_N(t)=R_t\Log_{x(t)}\wh x(t)$.
For every deterministic sequence of admissible recorded histories
$\mathcal D_N\in A_\Delta\cap E_D$, every subsequence admits a further
subsequence $N_j$ along which $m_{N_j} C_{\mathcal D_{N_j}}\to K$ uniformly
on $[0,T]^2$ and, in $C([0,T],\R^d)$,
\begin{equation}\label{cl:inf:subsequence-limit}
 \mathcal L\left(\sqrt{m_{N_j}}L_{N_j}\mid\mathcal D_{N_j}\right)
       \Rightarrow\mathcal L(\mathbb G_K),\quad
 \mathcal L\left(\sqrt{m_{N_j}}v_{N_j}\mid\mathcal D_{N_j}\right)
       \Rightarrow\mathcal L(\mathbb G_K).
\end{equation}
Here $\mathbb G_K$ is a continuous centered Gaussian
process with covariance $K(s,t)$. The same convergence holds along any
subsequence on which the normalized covariance converges uniformly.

\end{theorem}

For the tube in \eqref{cl:inf:tube}, let $\mathcal C_N$ denote successful
reporting and coverage of $x(t)$ at every $t\in[0,T]$. Under the random sampling design of Section~\ref{sec:active-strategy}, $\PP(\mathcal C_N)\longrightarrow1-\alpha$.
Thus every conditional weak limit of the normalized curve error is
Gaussian. Its covariance can depend on the recorded designs and the
subsequence; convergence of $mC_{\mathcal D_N}$ along the full sequence
gives a full sequence limit. These are conditional limits: averaging over
random designs need not give a Gaussian law. The feasible tube does not
require a unique limiting covariance, because the residuals consistently
estimate the covariance for the current design. Its radius has order
$m^{-1/2}$ in probability.

The coverage conclusion is uniform over model and design families with
the stated common bounds.
The pilot determines where to sample, while the final responses
estimate the curve and its uncertainty. The method assumes that actual
measurement positions are recorded; position error and physical
deployment costs are separate modeling questions.

\begin{algorithm}[H]
\caption{Active sampling and simultaneous tube}\label{alg:active}
\begin{algorithmic}[1]
\Require $N,n_0$, $m=N-n_0$, $x_0,T$, $b$, $J$, $\Delta=T/J$,
$k=\lceil\nu\rceil-1$, reconstruction and acceptance constants, and level $1-\alpha$.
\State {Fit \eqref{eq:estimator} to the $n_0$ pilot observations with bandwidth $b$, and integrate from $x_0$ to obtain $\gamma_0$.}
\State Construct pilot charts; collect $n_\ell\asymp m\Delta$ location--response pairs per cell,
sampling along and transverse to each pilot chord. Record cell labels and check chart admissibility and \eqref{cl:eq:design}.
\State Fit \eqref{cl:eq:fit} in coordinate velocities and check \eqref{cl:eq:coefficient-checks}.
\State Extend by \eqref{cl:eq:extended-field}, integrate \eqref{cl:eq:ode},
and compute residuals and \eqref{cl:eq:fitted-influence}.
\State Simulate the Gaussian process \eqref{cl:inf:feasible} with one Gaussian multiplier per response fixed across time.
Return $\wh x$ and \eqref{cl:inf:tube} using the simulated supremum quantile.
\end{algorithmic}
\end{algorithm}

On a chart, design, fit or integration failure, return
the prescribed constant curve and no tube; count the attempt as
noncoverage and count all allocated responses in the budget.
No step tests the unobservable event $A_\Delta$.

\section{Numerical results}\label{sec:numerics}

We compare passive kernel regression with pilot-guided neighborhood
sampling on $S^2$, $S^3$ and $\mathrm{SO}(3)$. The response budgets are
200, 400, 600, 800, 1000 and 2000, with 500 independent evaluation datasets
per setting and horizon $T=1$. Evaluation data are separate from
development data. The two methods use different sampling and calibration
rules, so these are comparisons at a common response count rather than
on identical observations. Additional torus and projective experiments
are reported in Appendices~\ref{app:extra-passive}--\ref{app:extra-active}.

For both passive and active methods, the kernel function is
\[
 K(u)=c_d\exp\!\left\{-\frac{1}{1-(\|u\|/0.9)^2}\right\}
       \mathbf1\{\|u\|<0.9\},
\]
where $c_d$ normalizes the kernel to integrate to one. The final active fits use quadratic local polynomials. In the active simulations the pilot uses ordinary kernel weights, giving
\begin{equation}\label{cl:eq:ordinary-pilot}
 \widehat V_b^{\rm ker}(x)=
 \frac{\sum_{i=1}^{n_0} K\bigl(\Log_x X_i/b\bigr)
                   \cP_{X_i\to x}Y_i}
      {\sum_{i=1}^{n_0} K\bigl(\Log_x X_i/b\bigr)}.
\end{equation}
Here the normalizing constant cancels. The implementation reports failure if the sum of the unnormalized weights is below $10^{-12}$.

Coverage is the proportion of all 500 attempts whose tube contains the
true curve at every evaluation node; a failed fit counts as noncoverage.
Bias in the tables is the average, over the 500 repetitions, of the maximum intrinsic error norm on the evaluation grid $\mathcal T$,
$500^{-1}\sum_{r=1}^{500}\max_{t\in\mathcal T}\|\Log_{x(t)}\widehat x^{(r)}(t)\|$,
including the prescribed constant-curve output after failure. Radius is the average tube radius over successfully constructed tubes. These are finite-grid
coverage summaries.

\begin{figure}[ht]
\centering
\includegraphics[width=\textwidth]{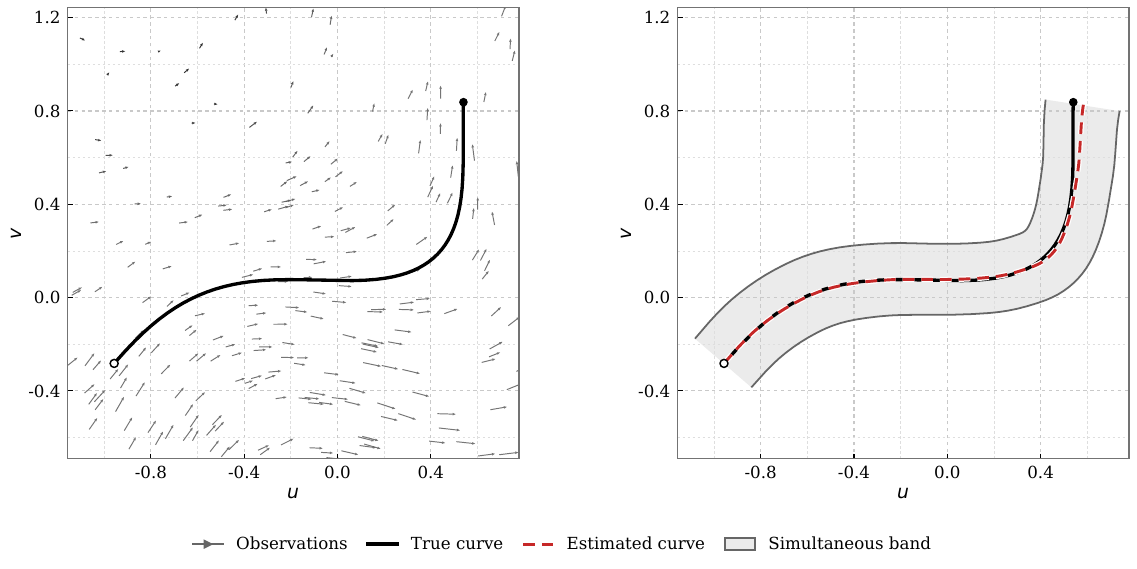}
\caption{Passive sampling on $S^2$. Left panel: noisy vector observations and the true curve. Right panel: the estimated curve and simultaneous confidence band, with the true curve for comparison.}
\label{fig:s2-passive-example}
\end{figure}

\begin{figure}[ht]
\centering
\includegraphics[width=\textwidth]{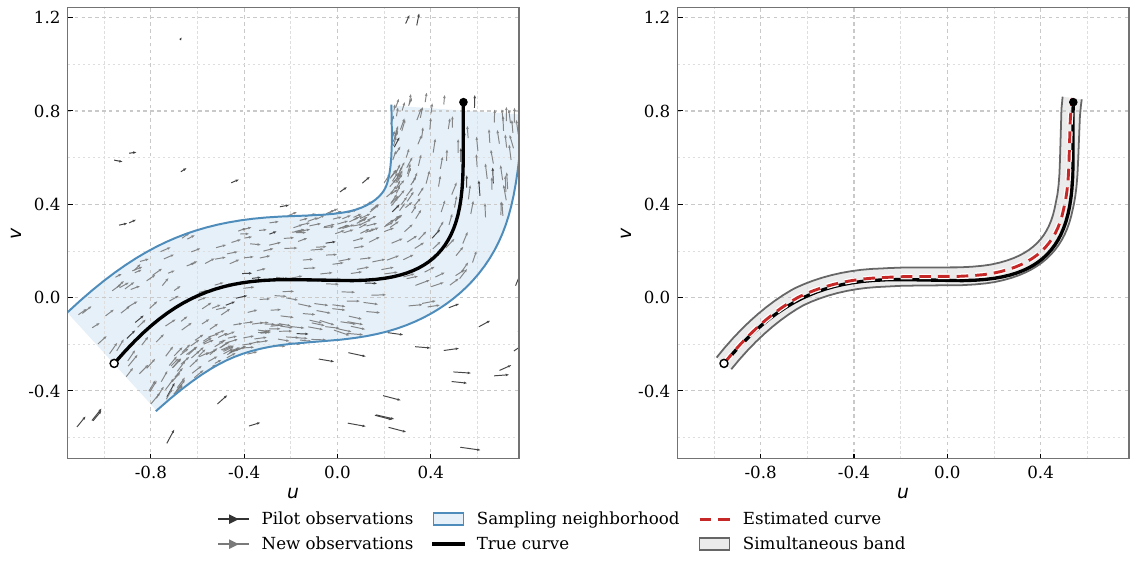}
\caption{Active sampling on $S^2$. Left panel: noisy vector observations and the true curve. Right panel: the estimated curve and simultaneous confidence band, with the true curve for comparison.}
\label{fig:s2-active-example}
\end{figure}

Figures~\ref{fig:s2-passive-example} and~\ref{fig:s2-active-example}
illustrate one evaluation dataset for each method at budget $N=1000$.
Both use an oblique Mercator chart centred at longitude
$1.1$ and latitude $0.4$ radians, with the map rotated $35^\circ$
counterclockwise.

\subsection{Passive kernel method}\label{app:passive-numerics}

The passive experiments implement Algorithm~\ref{alg:passive}, with the
uncorrected curve as the tube center. The fields include rotation with
changing latitude on $S^2$, rotation at unequal angular speeds on $S^3$,
and position-dependent angular velocity on $\mathrm{SO}(3)$.

Bandwidths were fixed before evaluation. All fits completed, so each mean radius uses 500 tubes.
We compute the curve, parallel frame and covariance factor using a Runge-Kutta solver with continuous
interpolation.
Two solves with successively tighter tolerances check numerical stability;
the second supplies the curve and factor used for Gaussian simulation.
All 15000 passive fits, including the additional manifolds in
Appendix~\ref{app:extra-passive}, passed these checks.
Appendix~\ref{app:passive-computation} gives the numerical settings.

{On the unit sphere $S^2$, the density relative to Riemannian volume is} $
 p(x)=(1+0.5x_1)/(4\pi), $ 
{and the field combines rotation with changing latitude, $
 V(x)=2.2(e_3\times x)
      +2.4(x_1^3-3x_1x_2^2)(e_3-x_3x). $}
The initial point is $x_0=(\sqrt3/2,0,1/2)$; {the reference curve is
computed by numerical integration of this field.}
Responses have independent tangent noise $0.2(I-xx^\top)Z$,
$Z\sim N(0,I_3)$. {The main bandwidth is $h=1.4n^{-2/9}$ at all six
sample sizes.}

On the unit sphere $S^3\subset\R^4$, use
\[
 p(x)=(1+0.5x_1)/(2\pi^2),
 V(x)=(-x_2,x_1,-0.35x_4,0.35x_3).
\]
The initial point is $x_0=(\sqrt3/2,0,1/2,0)$, and independent tangent
noise is $0.2(I-xx^\top)Z$ with $Z\sim N(0,I_4)$.
For all three-dimensional passive experiments, $h$ is the larger of
$4.5n^{-1/6}/\sqrt{\log n}$ and the bandwidth giving 12 expected kernel
neighbors at the minimum design density; the support radius is $0.9h$.

On $\mathrm{SO}(3)$ with the rotation-angle metric, use density $
 p(R)=(1+0.5R_{11})/(8\pi^2) $
and vector field with left-body angular velocity
$(1,0.35R_{11},0.25R_{21})$, starting at the identity. Independent
left-body noise has coordinate standard deviations
$(0.2(1+0.2R_{11}),0.15,0.25)$. Distances and radii are physical
rotation angles in radians. The main bandwidth follows the
three-dimensional rule above.

\begin{table}[ht]
\centering\small
\setlength{\tabcolsep}{2.5pt}
\caption{Passive sampling on three manifolds; $n$ is the sample size.}
\label{tab:passive-all}
\begin{tabular}{r rrr rrr rrr}
\toprule
& \multicolumn{3}{c}{$S^2$} & \multicolumn{3}{c}{$S^3$} & \multicolumn{3}{c}{$\mathrm{SO}(3)$} \\
\cmidrule(lr){2-4}\cmidrule(lr){5-7}\cmidrule(lr){8-10}
$n$ & Coverage (\%) & Bias & Radius & Coverage (\%) & Bias & Radius & Coverage (\%) & Bias & Radius \\
\midrule
200 & {95.2} & {0.156} & {0.390} & 89.6 & 0.103 & 0.158 & 87.8 & 0.127 & 0.188 \\
400 & {96.0} & {0.098} & {0.212} & 92.6 & 0.086 & 0.140 & 93.0 & 0.103 & 0.166 \\
600 & {96.6} & {0.080} & {0.166} & 93.8 & 0.078 & 0.131 & 93.0 & 0.093 & 0.157 \\
800 & {96.6} & {0.069} & {0.133} & 96.8 & 0.070 & 0.125 & 94.4 & 0.087 & 0.149 \\
1000 & {95.6} & {0.064} & {0.122} & 96.4 & 0.067 & 0.121 & 93.6 & 0.086 & 0.145 \\
2000 & {94.8} & {0.046} & {0.086} & 96.0 & 0.057 & 0.100 & 95.6 & 0.072 & 0.132 \\
\bottomrule
\end{tabular}
\end{table}

\subsection{Active neighborhood sampling}\label{sec:active-simulations}

{On all three manifolds, Algorithm~\ref{alg:active} uses the
same field, starting point and noise law as the corresponding passive
setting. The density $p$ describes the independent pilot locations.
We allocate $N/4$ responses to the pilot and $3N/4$ to the final batch.
The pilot bandwidth is $b=c(N/200)^{-1/6}$, with $c=0.9$, $1.8$ and
$2.4$ on $S^2$, $S^3$ and $\mathrm{SO}(3)$, respectively.}
Final observations are allocated as evenly as possible across time cells
and sampled uniformly along the pilot chords, with independent transverse
coordinates.
All acquired responses count toward $N$.
On $S^2$, use $J=(3,4,4,4,5,5)$, with independent uniform transverse
coordinates of half-widths $(0.4,0.8/3,0.8/3,0.8/3,0.8/3,0.2)$.
Calibration uses the 951st ordered maximum of 1000 Gaussian paths
on $\{j/1024:0\le j\le1024\}$, with each residual multiplier shared across time. Confidence tubes use the degrees-of-freedom correction in
\eqref{cl:inf:feasible}.

\begin{table}[ht]
\centering\small
\setlength{\tabcolsep}{2.5pt}
\caption{Active sampling; $N$ is the total sample budget.}
\label{tab:cloud-all}
\begin{tabular}{r rrr rrr rrr}
\toprule
& \multicolumn{3}{c}{$S^2$} & \multicolumn{3}{c}{$S^3$} & \multicolumn{3}{c}{$\mathrm{SO}(3)$} \\
\cmidrule(lr){2-4}\cmidrule(lr){5-7}\cmidrule(lr){8-10}
$N$ & Coverage (\%) & Bias & Radius & Coverage (\%) & Bias & Radius & Coverage (\%) & Bias & Radius \\
\midrule
200 & {94.0} & {0.078} & {0.139} & 97.2 & 0.118 & 0.237 & 97.6 & 0.091 & 0.172 \\
400 & {95.2} & {0.048} & {0.083} & 97.6 & 0.054 & 0.097 & 95.8 & 0.050 & 0.088 \\
600 & {95.2} & {0.032} & {0.060} & 96.8 & 0.061 & 0.115 & 97.8 & 0.053 & 0.098 \\
800 & {94.0} & {0.028} & {0.050} & 97.0 & 0.044 & 0.080 & 96.2 & 0.039 & 0.070 \\
1000 & {94.2} & {0.025} & {0.045} & 97.2 & 0.036 & 0.064 & 96.0 & 0.033 & 0.058 \\
2000 & {93.8} & {0.017} & {0.031} & 96.0 & 0.027 & 0.047 & 97.4 & 0.023 & 0.042 \\
\bottomrule
\end{tabular}
\end{table}

Across these three manifolds and six common budgets, the active mean radius
is smaller in most scenarios. The exception is $S^3$ at budget $N=200$,
where active and passive radii are $0.237$ and $0.158$.
At budget $N=400$ on $S^3$, active and passive coverage are $97.6\%$ and
$92.6\%$, with radii $0.097$ and $0.140$.
{On $S^2$ at budget 200, active and passive coverage are {$94.0\%$} and
$95.2\%$, with mean radii {$0.139$} and $0.390$. }
These comparisons evaluate both sampling and inference: the sampling
rules, quantiles and evaluation grids differ. The effect of sampling
alone therefore cannot be separated from the other differences.
\section{Real-data applications}\label{sec:appdesign}

\subsection{Regional ocean circulation on the sphere}\label{sec:drifter}

How much of a regional mean flow trajectory can existing ocean-drifter observations support? We study summer circulation in the North Atlantic box $[-55,-35]$ degrees longitude by $[25,40]$ degrees latitude. The Global Drifter Program hourly product \citep{elipot2016hourly} supplies 11977 eligible daily records from 221 buoys during June--August, 2010--2019. Entire buoys define training, validation and test sets of sizes 142, 33 and 46. The target is the cohort's average summer velocity, giving each buoy equal total weight, and its integral curve. Eastward and northward velocities are tangent vectors on $S^2$: the local constant estimator in \eqref{eq:estimator}, with these buoy weights, uses parallel transport to average them, and the fitted curve is integrated on the sphere.

\begin{figure}[!htbp]
\centering
\includegraphics[width=.58\textwidth]{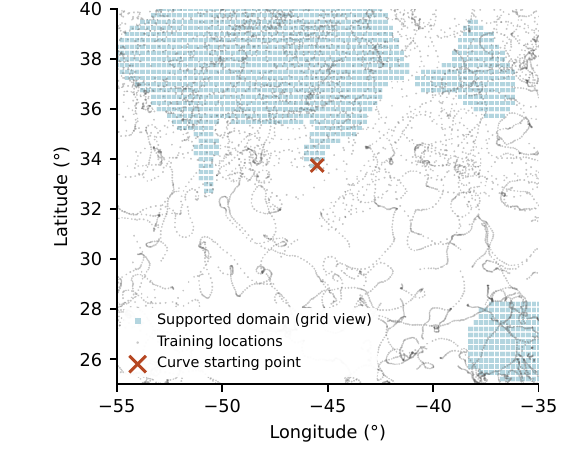}
\caption{North Atlantic drifters and spatial support. Daily training locations, the shaded supported domain and the selected starting point show where the mean flow route can be reconstructed.}
\label{fig:gdp-support}
\end{figure}

Spatial support limits the length of the reconstructed route.
Figure~\ref{fig:gdp-support} shows the training positions, the supported
domain $S$ at bandwidth 400 kilometres and the selected starting point.
The quarter-degree shading displays spatial support, which is checked
directly along the curve. We restrict reconstruction to a training-defined domain satisfying minimum distinct-buoy and effective-sample-size requirements in every training-fold omission. The original starting position $(-45,32.5)$ degrees is unsupported; its nearest admissible quarter-degree grid point is $(-45.5,33.75)$ degrees.
From this point, the estimated route remains supported for 3.79 days,
short of the intended five-day horizon. This illustrates the trade-off
between spatial resolution and observational support in drifter
climatologies \citep{laurindo2017improved}.
Appendix~\ref{app:drifter} gives the start selection, support
criteria, inclusion rules and temporal sensitivity analysis.

Within $S$, the fitted field reduces test velocity loss by $6.6\%$
relative to a transported constant-vector baseline, using 615 records
from 28 buoys. This improvement translates only weakly into prediction of individual drifter motion: endpoint errors are 24.40 versus 24.60 kilometres at one day and 127.91 versus 132.74 kilometres at five days. These comparisons involve only 17 and 12 completed test windows, respectively, each from four buoys; validation results show no consistent advantage.
Appendix~\ref{app:drifter} reports the errors and completed windows.
The curve describes regional mean flow, while individual motion also
contains temporal and eddy variability
\citep{lumpkin2013global,laurindo2017improved}. Dependence between
observations precludes interpreting the present analysis as a validation
of nominal confidence-tube coverage.

\subsection{Mean flow through a model canopy}\label{sec:lda}

How does the local mean flow route beside a canopy element change between positions within and above the canopy? Spatially averaged wind directions can conceal departures around individual elements. Using three dimensional velocimetry, \citet{harman2016spatial} found strong spatial variation in the mean flow within a model canopy, diminishing rapidly above it. We use active reconstruction to express this heterogeneity in terms of the lateral and vertical displacement of a finite mean streamline.

The public CSIRO Detailed Spatial Sampling archive \citep{harman2021data}
contains three-component laser Doppler anemometry (LDA) measurements at
1540 positions around a canopy of 60 mm-high elements. Each location was sampled for 35 seconds, and the released velocities are residence-time-weighted means. In a corridor
beside one element, 169 of 504 positions supply reference reconstructions
and 335 form the acquisition pool. The two mean streamlines start at the
same horizontal location and heights 42 and 84 mm; horizons of 10 and
5 milliseconds give similar downstream advances of about 26 mm. The target is an integral curve of the three-dimensional mean velocity, not an individual turbulent particle path.

Pointwise LDA permits measurement locations to be chosen, and adaptive acquisition has been demonstrated experimentally \citep{theunissen2015autonomous}. The present archive was collected on a prescribed grid, so we implement active selection retrospectively by revealing actual measurements at selected locations. A quarter of the total budget $N$ supplies a kernel pilot, which guides one further batch of observations to neighborhoods along the estimated route. The final reconstruction uses the unweighted local least-squares fits in \eqref{cl:eq:fit} and their integral curves. The main display uses $N=160$, including 40 pilot observations; Appendix~\ref{app:appdesign} gives the joint allocation rule and results for $N=80,160,240$, with 500 sampling repetitions at each height and budget.

\begin{figure}[!htbp]
\centering
\includegraphics[width=\textwidth]{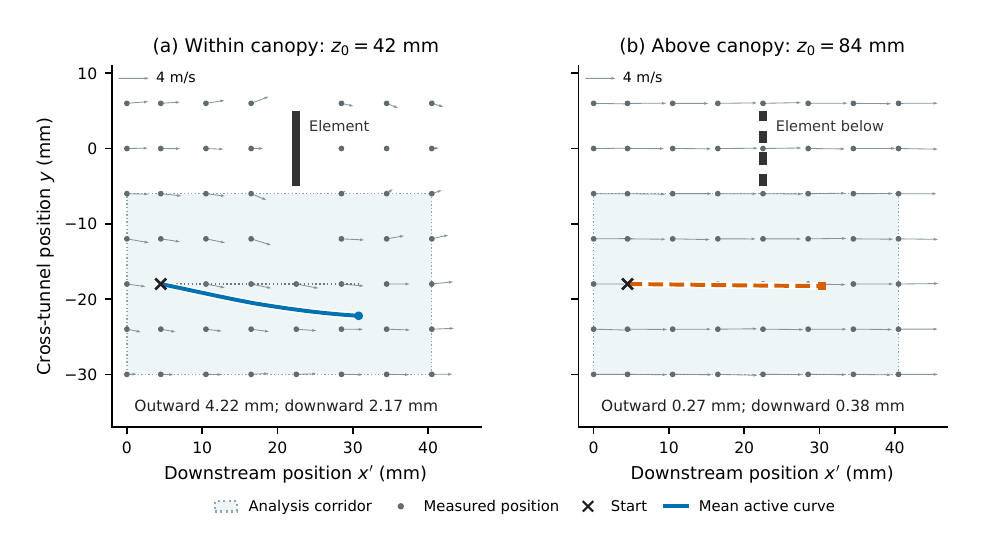}
\caption{Local mean flow reconstruction within and above a model canopy. Arrows show measured horizontal velocities, and curves show horizontal projections of the mean of 500 reconstructions.}
\label{fig:lda-canopy}
\end{figure}

Figure~\ref{fig:lda-canopy} shows horizontal projections of the mean of
500 three dimensional reconstructions, with measured horizontal
velocities on a common arrow scale. The paths may change height.
At $N=160$, mean outward displacement is 4.22 mm within the canopy and
0.27 mm above it; downward displacements are 2.17 and 0.38 mm.
Thus the within-canopy route bends outward and downward, whereas the
route above it remains nearly streamwise. The standard deviations across sampling designs are 0.19 and 0.06 mm for outward displacement, and 0.15 and 0.05 mm for downward displacement. These summarize variation in selected locations conditional on the archive. The height contrast also appears directly in the measurements: the median absolute lateral velocity in the corresponding local strips is 0.367 metres per second within the canopy and 0.039 metres per second above it.
Appendix~\ref{app:appdesign} gives the full results.

The greater within-canopy deflection gives a spatial interpretation of
the heterogeneity reported by \citet{harman2016spatial}. The contrast
persists across all three budgets and reference reconstruction orders. For the reserved observations, the quadratic reference gives outward displacements of 3.72 and 0.31 mm, whereas a linear reference gives 1.86 and 0.19 mm. The magnitude depends on reconstruction order. The selected locations
vary across repetitions, conditional on the archive; shared calibration
errors and unavailable raw time series prevent validation of nominal
confidence-tube coverage.

\section{Discussion}\label{sec:discussion}
Sampling design changes how accurately a fixed response budget can
recover an integral curve. Passive observations support reconstruction
and simultaneous inference; directing measurements toward a pilot can
improve accuracy under the stated localization and smoothness conditions.
The applications show how these curves describe local mean transport
and how observational support limits their reconstruction.

The applications suggest two extensions. First, repeated measurements
along a drifter track contain temporal dependence, and seasonal or eddy
variation can separate an individual trajectory from a mean flow route
\citep{lumpkin2013global,laurindo2017improved}. For dependent responses,
propagating the field error would also propagate cross-covariances between
observations; the present independent Gaussian multipliers would need to
be replaced by a construction that preserves that dependence. A
time-varying field would further require local information in time as
well as space, reducing the observations available near each part of the
route. These issues explain why improved mean-velocity prediction alone
does not establish coverage for individual drifter paths. Second, the
active method treats the response count as the sampling budget. Adaptive
LDA and autonomous underwater sampling show that directing measurements
is practical \citep{theunissen2015autonomous,kemna2018onboard}, but moving
between locations also consumes time. A design that balances localization
against travel cost would need to retain enough spread across the route
to identify transverse changes in the field. The canopy analysis shows a
related constraint even in a fixed archive: increasing the quota can
force observations farther from the pilot. Accounting for these costs
and spatial constraints would connect the statistical gain from active
sampling more directly to measurement effort.

\section*{Acknowledgment}
\par  Qirui Hu's research was supported by National Natural Science Foundation of China (NSFC) (Grant Nos.12601520), the Shanghai Engineering Research Center of Finance Intelligence (Grant No.~19DZ2254600) and by TRR 391 \textit{Spatio-temporal Statistics for the Transition of Energy and Transport} (Project number 520388526) funded by the Deutsche Forschungsgemeinschaft (DFG, German Research Foundation).
% BEGIN INCLUDED BIBLIOGRAPHY

% END INCLUDED BIBLIOGRAPHY

\clearpage
\appendix

\section{Numerical implementation details}\label{app:numerics}

\subsection{Passive calibration: residual pilots and forcing columns}
\label{app:passive-calibration-details}

This section specifies the auxiliary estimates used in
Section~\ref{sec:integrated}. Their consistency is established in the
supplementary proof of Theorem~\ref{thm:inference}.
Choose outer folds $F_1,\ldots,F_L$ by fixed observation indices or by
random labels independent of all observations. Put
$\mathcal I_\ell=\{1,\ldots,n\}\setminus F_\ell$ and partition each $\mathcal I_\ell$
into inner validation sets $J_{\ell f}$ in the same way. The numbers of
folds are fixed and all training and validation sizes are at least $cn$
for a fixed $c>0$. For an index set $I$, let $\wh V_{I,b}$ be
\eqref{eq:estimator} with sample $I$, denominator normalized by $|I|$,
and bandwidth $b$. For $i\in F_\ell$ and $j\in J_{\ell f}$, set
\[
 M_i=\wh V_{\mathcal I_\ell,b}(X_i),\qquad
 e_{\ell j}=Y_j-\wh V_{\mathcal I_\ell\setminus J_{\ell f},b}(X_j).
\]
Pool the inner residuals over $\mathcal I_\ell$ and define
\begin{equation}\label{arxiv-appendix:eq:nested-covariance}
 S_\ell^{\rm raw}(x)=
 \frac{|\mathcal I_\ell|^{-1}\sum_{j\in \mathcal I_\ell}W_g(x,X_j)
 (\cP_{X_j\to x}e_{\ell j})(\cP_{X_j\to x}e_{\ell j})^*}
 {s_\varpi\bigl(|\mathcal I_\ell|^{-1}\sum_{j\in \mathcal I_\ell}W_g(x,X_j)\bigr)}.
\end{equation}
Let $S_i$ be $S_\ell^{\rm raw}(X_i)$ with each eigenvalue replaced by
its maximum with $\gamma_n\downarrow0$. Fold-size-dependent thresholds
tending to zero are also allowed, as shown in the supplementary proof of
Theorem~\ref{thm:inference}.
The bandwidths obey \eqref{eq:pilotbw}. No derivative of the mean pilot
is required, because propagation uses the derivative of $\wh V_h$.

The following weights account for the smooth denominator stabilization.
At a query point $x$, put
\[
 \alpha_i(x)=\frac{W_h(x,X_i)}{n s_\varpi\bigl(\wh p_h(x)\bigr)},\qquad
 \bar M(x)=\sum_i\alpha_i(x)\cP_{X_i\to x}M_i,
\]
and define
\begin{align}
 \beta_i(x)&=s_\varpi'\bigl(\wh p_h(x)\bigr)\alpha_i(x)
  +\frac{1-s_\varpi'\bigl(\wh p_h(x)\bigr)\wh p_h(x)/s_\varpi\bigl(\wh p_h(x)\bigr)}{n},
  \label{eq:finite-denominator}\\
 D_i^{\rm pos}(x)&=\alpha_i(x)\cP_{X_i\to x}M_i-\beta_i(x)\bar M(x),&
 N_i(x)&=\alpha_i(x)\cP_{X_i\to x}S_i^{1/2}.
 \label{eq:finite-forcing}
\end{align}
Thus $\sum_iD_i^{\rm pos}(x)=0$ exactly. Where stabilization is inactive,
$\beta_i=\alpha_i$ and $D_i^{\rm pos}=\alpha_i(\cP M_i-\bar M)$.
The design and noise columns follow from perturbing observation weights
and responses, respectively, as verified below.

Let $\wh R_t$ be parallel transport along $\wh x$ to $T_{x_0}\M$, and set
$\wh A_t=\wh R_t(\nabla\wh V_h)_{\wh x_t}\wh R_t^{-1}$.
Draw mutually independent $\zeta_i\sim N(0,1)$ and standard Gaussian
$\eta_i\in T_{X_i}\M$, independently of the data. Use the same weights
at every time and solve the ordinary differential equation
\begin{equation}\label{arxiv-appendix:eq:finite-bootstrap}
 \dot Z_n^\#(t)=\wh A_tZ_n^\#(t)+
 \sum_{i=1}^n\wh R_t\bigl(D_i^{\rm pos}(\wh x_t)\zeta_i+N_i(\wh x_t)\eta_i\bigr),
 \qquad Z_n^\#(0)=0.
\end{equation}
Equation~\eqref{arxiv-appendix:eq:finite-bootstrap} has the same linear propagation
structure as the limiting SDE, with an estimated drift and Gaussian
forcing that retains the kernel overlap at bandwidth $h$. Conditional
on the Gaussian draws, it is an ordinary differential equation.
The process is in unscaled curve-error units, so its supremum quantile
can be used directly as a geodesic tube radius. In orthonormal
frames at the observations, concatenate the forcing columns into
$H_n(t)$ and solve $\dot U_n=\wh A U_n+H_n$, $U_n(0)=0$.
Then $Z_n^\#(t)=U_n(t)\xi$, with one standard Gaussian vector $\xi$, and
its entire conditional covariance kernel is
\begin{equation}\label{eq:finite-sandwich}
 \wh C_n(s,t)=U_n(s)U_n(t)^*.
\end{equation}
Changing an observation frame leaves this law unchanged.

To obtain the design-forcing columns in \eqref{eq:finite-forcing}, perturb the empirical observation weights by
$n^{-1}(\zeta_i-\bar\zeta)$ and differentiate the normalized kernel
ratio with responses $M_i$. Differentiating its numerator and
denominator gives precisely $\sum_iD_i^{\rm pos}\zeta_i$. Independent Gaussian
response perturbations with covariances $S_i$ give the columns $N_i$.

\subsection{Passive integration and refinement}\label{app:passive-computation}

The experiments use five outer folds for conditional means and five inner
folds for residual covariances. On $S^2$, auxiliary means use local linear
fits and covariances use smoothed residual squares. On $S^3$ and $\mathrm{SO}(3)$, the
auxiliary bandwidths are $(b,g_\Sigma)=(2.5,7)n^{-1/7}$ and
$(4,5)n^{-1/7}$, respectively.
Gaussian paths follow \eqref{arxiv-appendix:eq:finite-bootstrap}, with independent design
and response-noise weights across observations, shared across time.
We simulate 2000 paths on
$\mathcal T=\{j/1024:0\le j\le1024\}\cup\{0.1,0.9\}$.

The numerical settings aim to keep integration error small relative to
the statistical curve-error scale $a_n^{-1}$. Using the Runge-Kutta solver described
in Section~\ref{app:passive-numerics}, we limit every curve, frame, factor
and numerical-reference step to $\min(1,h^2)/16$. Coefficients are
evaluated at the integration times, retaining all nonzero kernel
contributions.

Stability is checked by comparing solutions at two tolerance levels.

\subsection{Active sampling and reconstruction}\label{app:active-settings}

Partition $[0,T]$ into $J$ equal time cells and allocate the final
observations as evenly as possible. On $S^3$ and $\mathrm{SO}(3)$, use
$J=(2,2,3,3,3,4)$ at the six increasing budgets. Final positions are
uniform along each pilot chord, with independent transverse coordinates
uniform on $[-0.4/J,0.4/J]$.
{On $S^2$, use $J=(3,4,4,4,5,5)$, with independent uniform transverse
coordinates of half-widths $(0.4,0.8/3,0.8/3,0.8/3,0.8/3,0.2)$.
}

For $S^3$ and $\mathrm{SO}(3)$, a position-only rule selects
the first admissible reconstruction window among four ordered pairs of
longitudinal margin and transverse half-width: $(0.5,0.6)$, $(0.75,1)$,
$(1,1.5)$ and $(1.5,2)$, each scaled by $1/J$. An observation retains
its acquisition cell label and enters its assigned fit only when its
recorded position lies in that cell's selected window. Excluded responses
still count toward $N$. 

Each cell is checked for sufficient sample size, Gram conditioning and
leverage; chart and fitted-field checks follow
Algorithm~\ref{alg:active}.

The active calibration uses the 951st ordered maximum among 1000 Gaussian
paths on $\{j/1024:0\le j\le1024\}$, with one residual multiplier per
response shared across time.

\subsection{Additional passive experiments}\label{app:extra-passive}

The torus experiments use the sample sizes, repetition count, kernel,
quantile rule and evaluation grid of Appendix~\ref{app:passive-computation}.
All fits completed. Distances are shortest
wrapped displacements in radians.

On $\mathbb T^2=\R^2/(2\pi\mathbb Z)^2$, use
\[
 p(\theta)=\frac{1+0.5\cos\theta_1}{(2\pi)^2},\qquad
 V(\theta)=(1,0.35\sin\theta_1),\qquad x_0=(0,0).
\]
The true curve has lift $x(t)=(t,0.35(1-\cos t))$.
Noise is $0.2\operatorname{diag}(1+0.2\sin\theta_1,0.75)Z$,
$Z\sim N(0,I_2)$, independently of locations.
The bandwidths are $h=3n^{-2/9}$, $b=2.5n^{-1/6}$ and
$g_\Sigma=7n^{-1/6}$. The constant in $h$ was selected from
$\{2,2.5,3\}$ by held-out prediction error on 24 independent development
datasets and fixed before validation.

On $\mathbb T^3=\R^3/(2\pi\mathbb Z)^3$, use
\[
 p(\theta)=\frac{1+0.5\cos\theta_1}{(2\pi)^3},\qquad
 V(\theta)=(1,0.35\sin\theta_1,0.35\cos\theta_1),\qquad x_0=0.
\]
Independent noise coordinates have standard deviations
$0.2(1+0.2\sin\theta_1)$, $0.15$ and $0.25$.
The main bandwidth follows the three-dimensional rule in
Section~\ref{app:passive-numerics}; $(b,g_\Sigma)=(5,6)n^{-1/7}$.

\begin{table}[htbp]
\centering\small
\setlength{\tabcolsep}{4pt}
\caption{Passive sampling on two tori; $n$ is the sample size.}
\label{tab:torus-finite-bandwidth-r33}\label{tab:torus3-finite-bandwidth-r37}
\begin{adjustbox}{max width=\linewidth}
\begin{tabular}{r rrr rrr}
\toprule
& \multicolumn{3}{c}{$\mathbb T^2$} & \multicolumn{3}{c}{$\mathbb T^3$} \\
\cmidrule(lr){2-4}\cmidrule(lr){5-7}
$n$ & Coverage (\%) & Bias & Radius & Coverage (\%) & Bias & Radius \\
\midrule
200 & 98.6 & 0.070 & 0.162 & 95.2 & 0.106 & 0.175 \\
400 & 96.6 & 0.057 & 0.120 & 94.4 & 0.096 & 0.163 \\
600 & 95.0 & 0.051 & 0.103 & 95.4 & 0.087 & 0.156 \\
800 & 96.6 & 0.044 & 0.091 & 97.4 & 0.081 & 0.151 \\
1000 & 96.4 & 0.040 & 0.083 & 95.0 & 0.081 & 0.148 \\
2000 & 96.6 & 0.032 & 0.065 & 96.0 & 0.075 & 0.138 \\
\bottomrule
\end{tabular}
\end{adjustbox}
\end{table}

The three-dimensional support floor is $O(n^{-1/3})$ and eventually
smaller than the logarithmic bandwidth, giving
$nh^6\sim4.5^6/(\log n)^3\to0$. The two-dimensional choices give
$nh^5=C^5n^{-1/9}\to0$, with $C=1.4$ on $S^2$ and $C=3$ on
$\mathbb T^2$, verifying the asymptotic undersmoothing condition.

The $\mathbb T^2$ result at $n=200$ combines high coverage with
59 local-support warnings. The smaller-budget undercoverage on $S^3$
and $\mathrm{SO}(3)$ is discussed with the main results.

\subsection{Additional active experiments}\label{app:extra-active}

The active torus and projective experiments follow the $S^3$ and
$\mathrm{SO}(3)$ sampling and reconstruction settings in
Appendix~\ref{app:active-settings}. The torus pilot densities, fields,
starting points and noise laws are those in Appendix~\ref{app:extra-passive}.
Their pilot bandwidth constants are $c=1.8$ for $\mathbb T^2$ and
$c=2.4$ for $\mathbb T^3$.

For $\mathbb{RP}^3$, represent a point by $[x]=\{x,-x\}$ on the unit
sphere $S^3$, with distance $\arccos|x^\top y|$. Relative to Riemannian
quotient volume, the pilot density is
\[
 p([x])=\frac{1+0.5x_1^2}{1.125\operatorname{vol}(\mathbb{RP}^3)}.
\]
The normalization follows from the uniform-sphere identity
$\mathbb E x_1^2=1/4$. The vector field has antipodally equivariant lift
$(-x_2,x_1,-0.35x_4,0.35x_3)$, starting at the $S^3$ initial point in
Section~\ref{app:passive-numerics}.
Tangent noise is isotropic Gaussian with standard deviation $0.2$.
The pilot bandwidth is $\min\{1.6,1.8(N/200)^{-1/6}\}$, keeping kernel
support inside the injectivity radius. The metric on $\mathbb{RP}^3$
differs from the physical rotation-angle metric on $\mathrm{SO}(3)$ by
a factor of two.

Each row has
500 successful tubes, with bias and radius as defined in
Section~\ref{sec:numerics}.

\begin{table}[htbp]
\centering\small
\setlength{\tabcolsep}{4pt}
\caption{Active sampling on two tori and $\mathbb{RP}^3$; $N$ is the total sample budget.}
\label{tab:cloud-extra}
\begin{adjustbox}{max width=\linewidth}
\begin{tabular}{r rrr rrr rrr}
\toprule
& \multicolumn{3}{c}{$\mathbb T^2$} & \multicolumn{3}{c}{$\mathbb T^3$} & \multicolumn{3}{c}{$\mathbb{RP}^3$} \\
\cmidrule(lr){2-4}\cmidrule(lr){5-7}\cmidrule(lr){8-10}
$N$ & Coverage (\%) & Bias & Radius & Coverage (\%) & Bias & Radius & Coverage (\%) & Bias & Radius \\
\midrule
200 & 95.6 & $0.035$ & 0.064 & 96.2 & 0.068 & 0.129 & 95.8 & 0.079 & 0.147 \\
400 & 94.2 & 0.024 & 0.043 & 97.0 & 0.042 & 0.074 & 97.2 & 0.052 & 0.093 \\
600 & 96.4 & $0.020$ & 0.036 & 96.8 & 0.041 & 0.075 & 96.6 & 0.057 & 0.104 \\
800 & 93.6 & 0.018 & 0.031 & 97.4 & 0.032 & 0.058 & 96.6 & 0.041 & 0.075 \\
1000 & 96.2 & $0.015$ & 0.027 & 95.8 & 0.028 & 0.049 & 95.6 & 0.035 & 0.060 \\
2000 & 95.2 & $0.011$ & 0.019 & 96.6 & 0.021 & 0.037 & 96.0 & 0.025 & 0.044 \\
\bottomrule
\end{tabular}
\end{adjustbox}
\end{table}

Across all five manifolds shared by the two methods, the active mean
radius is smaller in 29 of 30 settings; the exception is the $S^3$ setting
identified in Section~\ref{sec:active-simulations}. Across all six active
manifolds, mean radii are smaller at $N=2000$ than at $N=200$ and coverage
ranges from $93.6\%$ to $97.8\%$. Intermediate radii need not decrease
monotonically because the number of time cells changes with the budget.

\FloatBarrier

\section{Regional drifter analysis: cohort, support and completion}
\label{app:drifter}

\subsection{Data and supported reconstruction}

Drifter climatologies separate mean circulation from seasonal and eddy
variation \citep{lumpkin2013global}. Irregular sampling and correlated
observations make both spatial resolution and uncertainty sensitive to
how observations are pooled \citep{laurindo2017improved}. We therefore
split by buoy, weight buoys equally, and restrict reconstruction to
regions supported by several distinct buoys.

We use the Global Drifter Program hourly location, velocity and temperature data.\footnote{\href{https://doi.org/10.25921/x46c-3620}{GDP hourly data, doi:10.25921/x46c-3620}.} We retain 00:00 UTC records during June--August, 2010--2019 in the box of Section~\ref{sec:drifter}, with valid positions and velocities, interpolation gaps of at most 12 hours, and documented positive drogue-loss dates later than the observation. Zero or missing dates are excluded.

There are 11977 records and 221 eligible buoys: 7556 records from 142
training buoys, 1813 from 33 validation buoys and 2608 from 46 test buoys.
Buoys are assigned deterministically by identifier to training,
validation and test sets; the training folds use a separate identifier
hash.

For a fit containing $J$ eligible buoys, buoy $j$ with $n_j$ retained
records receives per-record weight $1/(Jn_j)$. Weights are recomputed for
the eligible training subset. Recorded eastward and northward velocities $(v_e,v_n)$ in metres per second are represented in their corresponding unit tangent directions by
\[
 Y=\frac{86400}{6371000}(v_e e_E+v_n e_N),
\]
in radians per day on the unit sphere, using Earth radius 6371000 metres. GDP velocity uncertainty fields are not substituted for the conditional covariance of the regression experiment. The comparisons use transported local constant regression, transported local linear regression and a common-anchor transported constant-vector baseline. The local constant method is the primary reconstruction in Section~\ref{sec:drifter}; the local linear fit checks sensitivity to a first-order spatial term.

\paragraph{Spatial support.}

None of the candidate bandwidths, 100, 200 and 400 kilometres, supports the original full-domain analysis in every training fold. Before evaluating held-out errors, we therefore fix the bandwidth at 400 kilometres and restrict reconstruction to the training-defined domain $S$. Its compact kernel neighborhoods have radius 360 kilometres. Membership requires at least ten distinct buoys, row effective sample size at least 30 and buoy effective sample size at least ten, both in the full training fit and in every training-fold omission. At a
query point, let $w_i$ include the fitting weight, volume correction and
kernel weight, and let $W_j$ be the sum for buoy $j$. The row and buoy
information summaries are
\[
 N_{\rm row}=\frac{(\sum_iw_i)^2}{\sum_iw_i^2},\qquad
 N_{\rm buoy}=\frac{(\sum_jW_j)^2}{\sum_jW_j^2}.
\]
A zero denominator is unsupported. These thresholds use locations and identifiers, without held-out responses.

The selected curve starting point is $(-45.5,33.75)$, obtained from
the search described in Section~\ref{sec:drifter}. The transported
constant-vector baseline retains the original reference point
$(-45,32.5)$: it averages responses transported there and transports
that mean to each query point.

Integration stops on leaving $S$, failing the one-bandwidth box margin or the land screen,\footnote{NOAA NCEI, ETOPO 2022 15 arc-second global relief model, \href{https://doi.org/10.25921/fd45-gt74}{doi:10.25921/fd45-gt74}.} or falling below speed $0.02$ metres per second. Each fitted subset must
also supply support within the original domain $S$. Intrinsic midpoint
integration is checked at a finer step size.

\subsection{Temporal sensitivity and prediction assessment}

We first examine how the reconstructed route changes when parts of the
observation period are omitted, then assess velocity and displacement
prediction on held-out buoys. The omissions probe the temporal composition
of this pooled mean-flow estimate; seasonal and interannual components
are treated explicitly in the climatology of \citet{lumpkin2013global}.

\begin{figure}[t]\centering
\includegraphics[width=\textwidth]{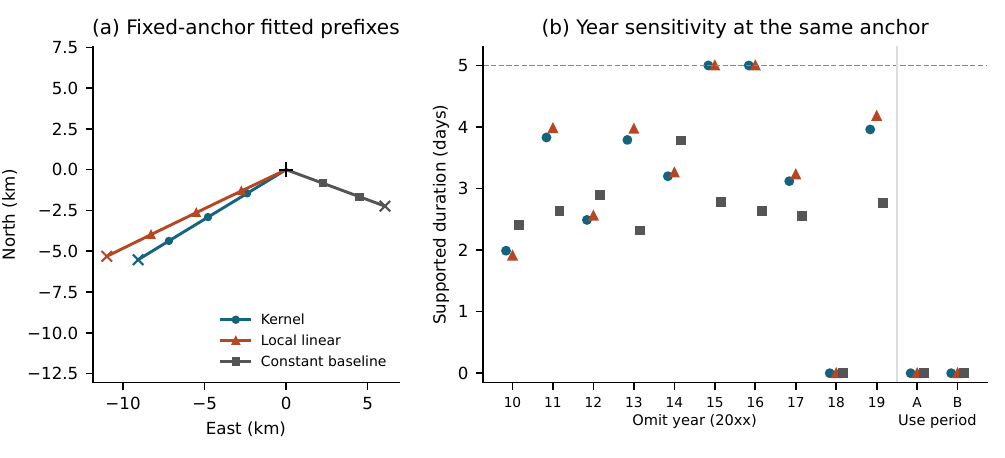}
\caption{Supported regional mean-flow reconstruction on $S^2$. (a) The three fitted prefixes in east/north normal coordinates at the starting point; markers denote complete days and crosses denote support stops. ``Kernel'' denotes transported local constant regression. (b) Supported durations after omitting each training year or retaining only 2010--2014 (A) or 2015--2019 (B), with the bandwidth, starting point and original support domain fixed. Zero denotes an unsupported start; five days denotes the intended horizon. Symbols identify the same estimators in both panels.}
\label{fig:gdp-curves}
\end{figure}

The primary local constant, local linear and baseline curves stop for
insufficient support after $3.79,3.97,2.69$ days. Each of ten
leave-one-year-out fits and the two half-period fits retains the fixed
bandwidth, anchor and original $S$, while also requiring support from its
own subset. All twelve subsets, including unsupported starts, are shown
in Table~\ref{tab:gdp-temporal} and Figure~\ref{fig:gdp-curves}.
Durations are in days; zero denotes an unsupported starting point.

\begin{table}[t]\centering\small
\setlength{\tabcolsep}{4pt}
\caption{Passive drifter observations on $S^2$: supported duration.}
\label{tab:gdp-temporal}
\begin{adjustbox}{max width=\linewidth}
\begin{tabular}{lrrr}
\toprule
Training subset & Local constant & Local linear & Baseline\\
\midrule
Omit 2010 & 1.99 & 1.91 & 2.41\\
Omit 2011 & 3.83 & 3.98 & 2.63\\
Omit 2012 & 2.49 & 2.56 & 2.89\\
Omit 2013 & 3.79 & 3.97 & 2.32\\
Omit 2014 & 3.20 & 3.26 & 3.78\\
Omit 2015 & 5.00 & 5.00 & 2.78\\
Omit 2016 & 5.00 & 5.00 & 2.64\\
Omit 2017 & 3.12 & 3.23 & 2.55\\
Omit 2018 & 0 & 0 & 0\\
Omit 2019 & 3.96 & 4.18 & 2.76\\
2010--2014 only & 0 & 0 & 0\\
2015--2019 only & 0 & 0 & 0\\
\bottomrule
\end{tabular}
\end{adjustbox}
\end{table}

\paragraph{Held-out prediction.}

Within the supported region, both local regressions improve held-out velocity prediction over the constant-vector baseline (Table~\ref{tab:gdp-velocity}). Squared vector errors are averaged first within buoy and then across eligible buoys. Local constant regression has the lowest validation loss; on test data the two local losses are nearly equal and about $6.6\%$--$6.7\%$ below the baseline. This suggests that spatially varying mean velocity is useful for prediction in the sampled region, while adding a local linear term yields little further test improvement at this bandwidth. The result concerns 615 supported test records from 28 buoys, out of 2608 records from 46 test buoys.

\begin{table}[t]
\centering\small
\caption{Passive drifter observations on $S^2$: velocity prediction, with losses in $10^{-3}\,\mathrm{m}^2\mathrm{s}^{-2}$.}
\label{tab:gdp-velocity}
\begin{adjustbox}{max width=\linewidth}
\begin{tabular}{lrrrrr}
\toprule
Split & Local constant & Local linear & Baseline & Supported records/buoys & All \\
\midrule
Validation & 148.635 & 152.143 & 152.283 & 735/26 & 1813/33 \\
Test       & 146.954 & 146.761 & 157.313 & 615/28 & 2608/46 \\
\bottomrule
\end{tabular}
\end{adjustbox}
\end{table}

For displacement prediction, a window consists of six consecutive daily
records spanning five days. A greedy chronological rule accepts the next
window from a buoy only when its start is at least five days after the
preceding start. Time interiors are disjoint, although neighboring
windows may share an endpoint. The 811 windows comprise 334 from 30
validation buoys and 477 from 43 test buoys. Of these, 131 validation
windows from 24 buoys and 113 test windows from 25 buoys start in $S$.

Completion requires reaching the horizon at both numerical step sizes,
with endpoint discrepancies at most one metre. A common comparison
requires all three methods to complete the same window. Only 17 test
windows from four buoys qualify at one day, and 12 from four buoys at
five days. Table~\ref{tab:gdp-displacement} reports geodesic endpoint
errors in kilometres, averaged first within buoy and then across buoys
in each common completion subset. Gains over the baseline are smaller
and less consistent than for velocity prediction.
Table~\ref{tab:gdp-own} records method-specific and common completion
counts; Table~\ref{tab:gdp-stops} accounts for every window, including
those excluded from the endpoint-error comparison.

\begin{table}[t]
\centering\small
\caption{Passive drifter observations on $S^2$: displacement prediction.}
\label{tab:gdp-displacement}
\begin{adjustbox}{max width=\linewidth}
\begin{tabular}{llrrrr}
\toprule
Split & Horizon & Windows/buoys & Local constant & Local linear & Baseline \\
\midrule
Validation & 1 day  & 14/4 & 38.05 & 38.39 & 36.51 \\
Validation & 5 days & 10/3 & 125.34 & 122.93 & 123.41 \\
Test       & 1 day  & 17/4 & 24.40 & 24.58 & 24.60 \\
Test       & 5 days & 12/4 & 127.91 & 129.34 & 132.74 \\
\bottomrule
\end{tabular}
\end{adjustbox}
\end{table}

\begin{table}[t]\centering\small
\setlength{\tabcolsep}{4pt}
\caption{Passive drifter observations on $S^2$: numerical completion.}
\label{tab:gdp-own}
\begin{adjustbox}{max width=\linewidth}
\begin{tabular}{llrrrr}
\toprule
Split & Horizon & Local constant & Local linear & Baseline & Common\\
\midrule
Validation & 1 day & 14/4 & 14/4 & 15/5 & 14/4\\
Validation & 5 days & 10/3 & 12/4 & 14/4 & 10/3\\
Test & 1 day & 17/4 & 17/4 & 18/4 & 17/4\\
Test & 5 days & 13/4 & 13/4 & 15/4 & 12/4\\
\bottomrule
\end{tabular}
\end{adjustbox}
\end{table}

Table~\ref{tab:gdp-stops} lists the five-day stopping reasons for all
windows. Each row sums to all windows in its split, with categories
defined by the algorithm's stopping conditions.

\begin{table}[t]\centering\small
\setlength{\tabcolsep}{4pt}
\caption{Passive drifter observations on $S^2$: stopping reasons.}
\label{tab:gdp-stops}
\begin{adjustbox}{max width=\linewidth}
\begin{tabular}{llrrrrrr}
\toprule
Split & Method & Box margin & Horizon & Support & Low speed & Unresolved crossing & Total\\
\midrule
Validation & Local constant & 266 & 10 & 56 & 1 & 1 & 334\\
Validation & Local linear   & 266 & 12 & 55 & 1 & 0 & 334\\
Validation & Baseline       & 266 & 14 & 54 & 0 & 0 & 334\\
Test & Local constant & 340 & 13 & 122 & 1 & 1 & 477\\
Test & Local linear   & 340 & 13 & 122 & 1 & 1 & 477\\
Test & Baseline       & 340 & 15 & 122 & 0 & 0 & 477\\
\bottomrule
\end{tabular}
\end{adjustbox}
\end{table}

Only $17/477=3.56\%$ of all test windows enter the common one-day comparison
and $12/477=2.52\%$ the common five-day comparison. Both proportions use all test windows as the denominator, whereas
method-specific completion and support-restricted proportions use different
denominators. The endpoint errors therefore describe only the small subset
on which every method completes.

\FloatBarrier

\section{Canopy LDA analysis: active design and sensitivity}\label{app:appdesign}

\subsection{Data and active reconstruction}

The strong within-canopy spatial variation documented by
\citet{harman2016spatial} motivates reconstructing local routes at two
heights, rather than summarizing the flow by one prevailing direction.
The observations are version 1 of the CSIRO Detailed Spatial Sampling collection, \href{https://doi.org/10.25919/xs2b-be57}{doi:10.25919/xs2b-be57}. The experiment and processing are described by \citet{harman2016spatial} and the collection's data overview. The archive provides three mean velocities and turbulent moments, but no raw time series or formal uncertainties for the means. Turbulent moments are not standard errors of those means. The overview identifies systematic and dependent error sources associated with the common instrument setup. Each measured position represents 35 seconds of acquisition, excluding repositioning and calibration.

Let $x'=x-2928.5$ mm, where $x$ is the source coordinate. We use $x'\in[0,40.5]$, $y\in[-30,-6]$ and $z\in[24,108]$ mm, beside the central 10 mm-wide canopy element and away from its solid interior. The starts are $(4.5,-18,42)$ and $(4.5,-18,84)$ mm, with horizons $T=0.010$ and $0.005$ seconds. Recorded velocities in metres per second are multiplied by 1000 for integration in millimetres. All three velocity components are retained. Outward displacement is $-18-\widehat y(T)$, downward displacement is $z_0-\widehat z(T)$, and downstream advance is $\widehat x'(T)-4.5$, all in millimetres.

Source indices $i_x=0,\ldots,7$, $i_y=0,\ldots,9$ and $i_z=0,\ldots,19$ identify the eight downstream sections, ten lateral positions and heights in descending order. The condition $(i_x+i_y+i_z)\bmod3=0$ reserves 169 corridor observations; the remaining 335 form the acquisition pool. Missing positions are not imputed. Reference observations enter neither the pilot nor the final budgeted fits or selection decisions. The subsets share experimental calibration and conditions.

\paragraph{Acquisition from the measured pool.}
Adaptive LDA can allocate measurements according to local curvature,
statistical error and sample spacing \citep{theunissen2015autonomous}.
Here the allocation targets neighborhoods along a pilot route in the
existing archive.
For each $N\in\{80,160,240\}$, a uniform sample of $n_0=N/4$ distinct pool positions supplies a kernel pilot. The smooth compact radial kernel is $K(r)\propto\exp\{-1/(1-r^2)\}\mathbf1\{r<1\}$ with radius 36 mm. Pilot positions at times 0 and $T/2$ define two anchors, and each segment receives $q=(N-n_0)/2$ final responses. The final locations are chosen using positions and pilot responses alone; their responses are revealed together after selection.

To allocate distinct points jointly, let $A$ and $B$ be the unqueried candidate sets within a common radius $R$ of the two anchors. Quotas of size $q$ are feasible when $|A|\ge q$, $|B|\ge q$ and $|A\cup B|\ge2q$. We start at $R=40$ mm and, when necessary, choose the smallest radius satisfying these conditions among actual candidate distances. For candidate $i$ and segment $j$, let $r_{ij}$ be its distance in millimetres to the anchor. With independent standard Gumbel variables $G_{ij}$, we minimize the total assignment cost $r_{ij}^2/(2\cdot12^2)-G_{ij}$ subject to the two quotas and no reused point. Thus overlapping neighborhoods are allocated simultaneously, with a preference for nearby positions. The spatial costs govern acquisition only.

For each segment, the ten monomials of total degree at most two are fitted by the unweighted criterion \eqref{cl:eq:fit}, separately for the three velocity components and using only that segment's assigned final responses. Coordinates are centered at the anchor and scaled by 20 mm for numerical conditioning. This rescaling leaves the polynomial space and fitted field unchanged. In Euclidean space the exponential-map differential is the identity. The two fitted fields are integrated successively as in \eqref{cl:eq:ode}. Every pilot and final observation counts toward $N$; the nominal acquisition times are 46.7, 93.3 and 140 minutes at the three budgets.

The finite grid entails sampling without replacement and joint allocation, rather than the independent continuous-location experiment of Section~\ref{sec:cloud}. Together with the shared errors of archived means, this distinction makes the exercise a retrospective assessment of reconstructed routes, without a claim of theoretical rate or confidence-coverage validation. Each budget and height has 500 sampling repetitions, with parameters fixed after separate development runs. The ODE uses fourth-order Runge-Kutta on 100 steps. Infeasible allocation, fewer than 14 regression rows, a design singular-value condition number above $10^5$, or integration leaving the corridor returns the initial constant curve, which remains in every summary. Failed pilots likewise return a constant pilot. All 3000 active runs succeeded. Above the canopy at $N=240$, four runs require radius expansion, to at most 40.67 mm; no other setting requires expansion. Assignment, distinct budgets, deterministic replay and numerical integration were checked.

\subsection{Route features and reconstruction sensitivity}
\begin{table}[!htbp]\centering\small
\setlength{\tabcolsep}{5pt}
\caption{Endpoint displacements of active mean-flow reconstructions (mm): mean over 500 sampling designs, with the between-design standard deviation in parentheses. The standard deviations describe location-selection variability conditional on the archive, not measurement-error confidence intervals.}
\label{tab:lda-displacements}
\begin{adjustbox}{max width=\linewidth}
\begin{tabular}{llrrr}\toprule
Flow & $N$ & Downstream & Outward & Downward \\
\midrule
Within canopy & 80 & 26.08 (0.55) & 4.26 (0.30) & 2.13 (0.22) \\
Within canopy & 160 & 26.29 (0.36) & 4.22 (0.19) & 2.17 (0.15) \\
Within canopy & 240 & 26.29 (0.27) & 3.75 (0.16) & 2.02 (0.12) \\
Above canopy & 80 & 25.79 (0.21) & 0.21 (0.08) & 0.39 (0.06) \\
Above canopy & 160 & 25.82 (0.17) & 0.27 (0.06) & 0.38 (0.05) \\
Above canopy & 240 & 25.99 (0.16) & 0.44 (0.07) & 0.41 (0.05) \\
\bottomrule\end{tabular}
\end{adjustbox}\end{table}

Table~\ref{tab:lda-displacements} shows that the stronger outward and downward departure within the canopy persists at all budgets. Mean outward displacement ranges from 3.75 to 4.26 mm within the canopy and from 0.21 to 0.44 mm above it, over similar downstream advances. The main figure averages the 500 active reconstructed paths at each common time and plots the resulting mean downstream and transverse coordinates. It summarizes the reconstructions; it is not a separately integrated velocity field or a confidence tube. The displayed $N=160$ setting was fixed before the formal runs. Shading marks the horizontal projection of the analysis corridor. The bar represents the canopy element with schematic thickness and is dashed when the element lies below the displayed height. Dots mark measured positions, the cross marks the common horizontal start, and dotted horizontal segments mark the initial streamwise direction.

At each query position, the reference fits a total-degree-two polynomial to the reserved observations using Gaussian distance weights of scale 15 mm, then integrates the fitted local velocity. Sensitivity reconstructions use scales 12 and 18 mm, the complementary acquisition subset at scale 15 mm, and a local linear fit at scale 15 mm. These are checks on reconstructed route features, not additional acquisition methods or observed curve truth. The primary reference paths have positive streamwise velocity throughout and remain within 4.28 and 3.03 mm of actual measurement locations. Direct velocity summaries use $x'\in[4.5,34.5]$ and $y\in[-24,-12]$, with heights 36--48 mm within the canopy and 78--96 mm above it. These strips contain 50 and 54 measured positions, respectively.

The reserved-data quadratic reference gives outward and downward displacements of 3.72 and 2.09 mm within the canopy, versus 0.31 and 0.40 mm above it. Alternative quadratic references differ from these paths by at most 0.33 and 0.12 mm. The linear reference changes the within-canopy path by as much as 2.42 mm and gives outward displacements of 1.86 and 0.19 mm. Thus the direction and height contrast are more stable than the displacement magnitudes.

\begin{table}[!htbp]\centering\small
\setlength{\tabcolsep}{4pt}
\caption{Mean maximum three-dimensional disagreement between the active reconstructions and alternative reference curves (mm), evaluated at 101 times. Quadratic reference scales are indicated; the complementary subset and linear reference use scale 15 mm.}
\label{tab:lda-reference-sensitivity}
\begin{adjustbox}{max width=\linewidth}
\begin{tabular}{llrrrrr}\toprule
Flow & $N$ & 15 mm & 12 mm & 18 mm & Complementary & Linear \\
\midrule
Within canopy & 80 & 0.882 & 0.767 & 1.055 & 0.872 & 2.784 \\
Within canopy & 160 & 0.666 & 0.554 & 0.837 & 0.604 & 2.812 \\
Within canopy & 240 & 0.375 & 0.392 & 0.498 & 0.506 & 2.372 \\
Above canopy & 80 & 0.285 & 0.272 & 0.366 & 0.246 & 0.907 \\
Above canopy & 160 & 0.257 & 0.265 & 0.336 & 0.196 & 0.942 \\
Above canopy & 240 & 0.411 & 0.462 & 0.446 & 0.313 & 1.133 \\
\bottomrule\end{tabular}
\end{adjustbox}\end{table}

The linear-reference discrepancies in Table~\ref{tab:lda-reference-sensitivity} are larger than those from changing the quadratic scale. These differences reflect reconstruction dependence and are not calibrated errors against physical truth. Finite spatial support also matters: above the canopy, the mean distance of selected locations from their assigned anchors increases from 17.13 to 19.68 and 23.99 mm as the budget grows. Disagreement with the primary reference decreases from 0.285 to 0.257 mm and then rises to 0.411 mm. A larger quota in a fixed archive therefore need not improve localization, even when allocation is feasible. The height contrast remains present despite this variation.

\end{document}